\documentclass[letterpaper, 10 pt, conference]{ieeeconf}

\IEEEoverridecommandlockouts
\usepackage{graphicx}
\usepackage{amsmath,amssymb}
\usepackage{booktabs}
\usepackage{multirow}
\usepackage{array}
\usepackage{cite}
\usepackage{url}
\usepackage{xcolor}
\usepackage{booktabs}
\usepackage{makecell}
\usepackage[table]{xcolor}
\usepackage{multirow}

\usepackage{amsmath}
\usepackage{amssymb}
\usepackage{amsfonts}
\usepackage{bm}
\usepackage{mathtools}

\usepackage{graphicx}
\usepackage{float}
\usepackage{capt-of}

\usepackage{booktabs}
\usepackage{multirow}
\usepackage{array}
\usepackage{tabularx}

\usepackage{xcolor}
\usepackage[most]{tcolorbox}

\usepackage{cite}
\usepackage{url}
\usepackage[
    colorlinks=true,
    linkcolor=blue,
    citecolor=blue,
    urlcolor=blue
]{hyperref}

\usepackage[T1]{fontenc}
\usepackage[most]{tcolorbox}
\usepackage{listings}

\tcbuselibrary{listings,breakable}

\newtcblisting{promptbox}[1]{
    enhanced,
    breakable,
    colback=gray!3,
    colframe=black!55,
    boxrule=0.5pt,
    arc=1pt,
    left=5pt,
    right=5pt,
    top=5pt,
    bottom=5pt,
    title=\textbf{#1},
    fonttitle=\small,
    listing only,
    listing options={
        basicstyle=\ttfamily\scriptsize,
        breaklines=true,
        columns=fullflexible,
        keepspaces=true,
        showstringspaces=false,
        escapeinside={(*@}{@*)}
    }
}
\definecolor{MyGreen}{rgb}{0.4, 0.9, 0.1}

\title{CARE: Experience-Guided Atomic Corrective Execution for Vision-Language-Action Policies}

\author{
Junlan Xiao$^{1*}$,
Junwei Jiang$^{1*}$,
Zaibin Zhang$^{1*}$,
Yifan Wang$^{1}$,
Zhongbo Zhang$^{1}$,
Huchuan Lu$^{1}$,
Lijun Wang$^{1\dagger}$%
\thanks{$^{1}$All authors are with Dalian University of Technology,
Dalian, China.}%
\thanks{$^{*}$Equal contribution.}%
\thanks{$^{\dagger}$Corresponding author.}%
}

\IEEEaftertitletext{%
\begin{minipage}{\textwidth}
    \centering
    \includegraphics[width=\textwidth]{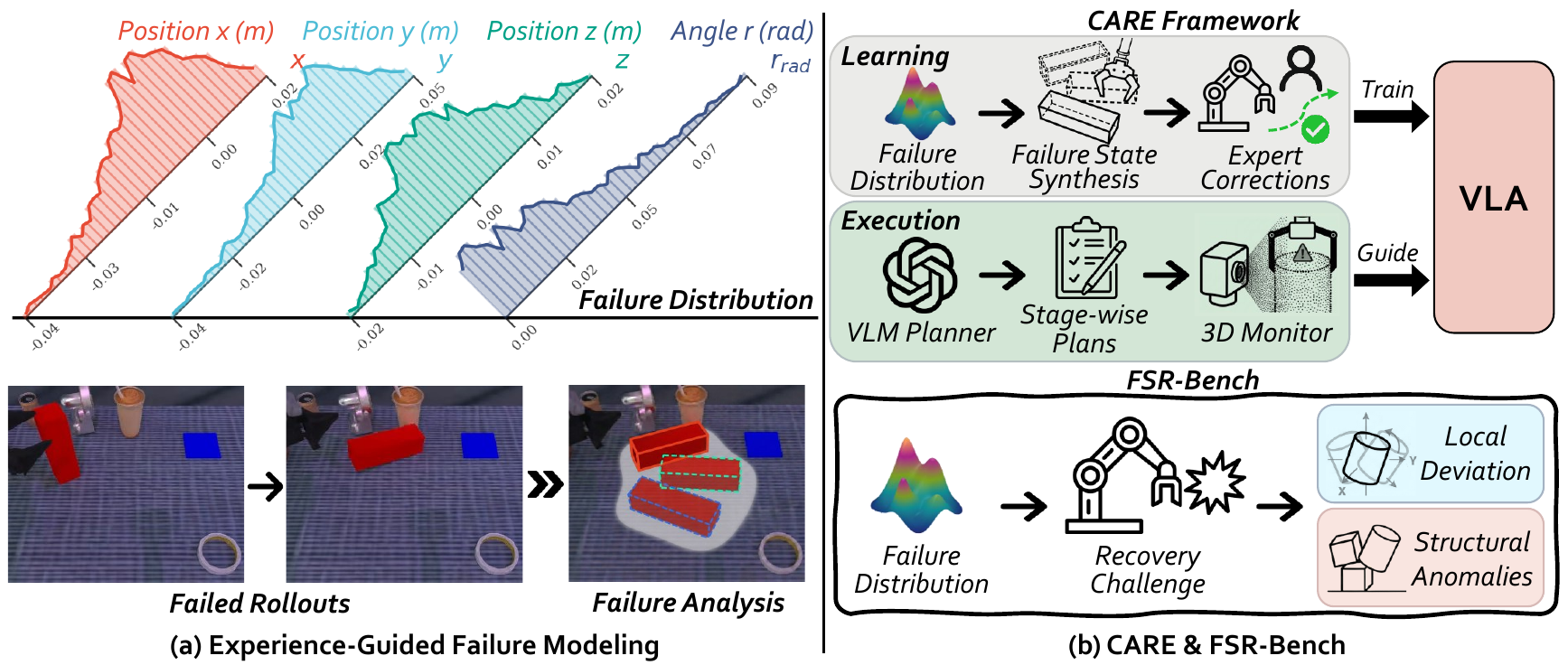}
    \captionof{figure}{
    Overview of CARE and FSR-Bench.
    (a) CARE models stage-specific failure distributions from execution
    experience to generate failure states and collect corrective demonstrations.
    (b) CARE uses these distributions for recovery learning, whereas FSR-Bench
    independently estimates its evaluation distribution from a mixture of
    nominal policies to generate shared recovery initializations.
}
    \label{fig:teaser}
\end{minipage}
\vspace{0.5\baselineskip}%
}

\begin{document}

\maketitle
\thispagestyle{empty}
\pagestyle{empty}

\begin{abstract}
Vision-Language-Action~(VLA) policies achieve strong performance in robotic manipulation but remain brittle once execution deviates from nominal trajectories.
We propose CARE~(Corrective Atomic Robotic Execution), a framework that improves recovery by learning from failures encountered during execution.
Instead of generating corrective data from manually designed or random perturbations, CARE collects failed rollouts, models stage-conditioned post-failure deviations, and uses the resulting empirical distributions to synthesize representative failure states and corrective demonstrations.
At inference time, CARE combines stage-wise planning with physically grounded 3D monitoring to trigger atomic adjustments or re-operations while preserving task progress.
We further introduce the Failure State Recovery Benchmark~(FSR-Bench), which evaluates recovery from intermediate failure states under local deviations and structural anomalies.
Experiments across multiple VLA backbones, simulation benchmarks, and real-world dual-arm tasks show consistent improvements, with average task-success gains of 14.5 points in simulation and 15.9 points in the real world. Code, models, and data are available at \url{https://github.com/xiaojunlan/care}
\end{abstract}

\section{Introduction}
\label{sec:introduction}

Vision-Language-Action~(VLA) policies have substantially improved the generality of robotic manipulation~\cite{black2024pi_0,pertsch2025fast}. 
However, most policies are still trained primarily on successful expert demonstrations and evaluated from nominal initial states~\cite{liu2023libero,james2020rlbench,qin2025robofactory,chen2025robotwin}.
This leaves an important gap between training and deployment: once execution deviates from the nominal trajectory, the policy may enter states that are rarely represented in its training data.
Such deviations are common in physical manipulation, e.g., objects may slip, tip, or be misplaced, and coordinated arms may become unsynchronized. All these deviations can quickly propagate into task failure.
Thus, robust manipulation requires not only executing nominal behaviors, but also recovering from the intermediate failure states induced by execution itself.

Existing recovery methods mainly address this problem in two ways.
One line of work detects failures and rolls the system back or replans from an earlier state~\cite{pan2025self,xu2025affordance}.
While effective in some settings, these methods largely treat failures as exceptions to be removed rather than experience that can improve the policy.
Another line of work trains recovery behaviors from corrective demonstrations collected around perturbed states~\cite{lin2025failsafe,dai2025racer}.
However, generic perturbation strategies do not explicitly model which post-failure deviations are actually induced by policy execution.
This motivates a different question:
\emph{can execution failures themselves provide structured supervision for learning recovery?}

We answer this question with CARE (\textbf{C}orrective \textbf{A}tomic \textbf{R}obotic \textbf{E}xecution), a framework that turns execution experience into corrective supervision and uses it for closed-loop recovery.
CARE first collects failed rollouts and characterizes stage-conditioned geometric deviations, such as translational and rotational errors.
These empirical failure distributions are then used to synthesize representative failure states and collect targeted corrective demonstrations.
Instead of training only on nominal trajectories or arbitrary perturbations, the VLA is therefore exposed to recovery behaviors around failure states that are grounded in its own execution experience.

Learning corrective behaviors alone is insufficient if the system does not know when and how to invoke them.
CARE therefore organizes long-horizon manipulation into atomic stages and uses a physically grounded 3D monitor to track stage-relevant geometric conditions during execution.
When a deviation is detected, the system issues either an intra-execution adjustment for local correction or a post-execution re-operation when the current atomic stage must be repeated.
The same VLA executor handles both nominal and corrective instructions, allowing recovery without switching to a separate recovery policy.

Evaluating such behavior is also difficult with existing manipulation benchmarks, which primarily measure task success from nominal initial states.
We therefore introduce the Failure State Recovery Benchmark~(FSR-Bench), which evaluates recovery directly from intermediate failure states.
FSR-Bench contains local geometric deviations as well as structural failures that require multi-step correction or coordinated multi-arm recovery, and reports recovery success independently from nominal task execution.

We evaluate CARE on RoboTwin~2.0~\cite{chen2025robotwin},
RoboFactory~\cite{qin2025robofactory}, FSR-Bench, and real-world
dual-arm manipulation tasks. Across benchmark--backbone settings,
CARE yields an average task-success gain of 14.5 points
in simulation and 15.9 points in the real world, together with a
7.5-point gain in recovery success on FSR-Bench. These
results demonstrate that execution failures provide effective
supervision for learning corrective behaviors in VLA policies.

\section{Related Work}
\subsection{Robotic Self-correction Systems}

Robust manipulation relies on the synergy between error detection and recovery. For detection, early methods use multimodal fusion~\cite{xu2025can,bu2024closed,li2018slip} or human intervention~\cite{shi2024yell}, while recent VLM-based approaches~\cite{xia2025phoenix,duan2024aha,dai2025racer,qi2026self,pan2025omnimanip} may lack physical grounding and be prone to misjudgment. World models~\cite{pan2025self} can incur latency and hallucination, while 3D point cloud-based monitoring~\cite{zhou2025code} improves physical grounding but remains limited to single-arm settings. For recovery, prior methods mainly rely on rollback~\cite{pan2025self,xu2025affordance}, VLM-based correction~\cite{yang2025fpc}, or corrective supervision from random perturbations~\cite{lin2025failsafe,dai2025racer} and human interventions~\cite{intelligence2025pi}. RL-based methods can also learn robust recovery behaviors~\cite{luo2024serl,vats2023efficient}, but real-robot RL may require reward specification, online exploration, and environment resets. Existing approaches, however, do not explicitly model stage-conditioned post-failure deviations to guide corrective data synthesis. CARE models this empirical failure structure to learn and invoke atomic corrections in multi-arm manipulation.


\subsection{Vision-Language-Action Models}
Vision-Language-Action models have emerged as a general paradigm for language-conditioned robot manipulation.
Existing models mainly generate actions through continuous generative modeling, such as diffusion or flow matching~\cite{black2024pi_0,liu2024rdt,wen2025dexvla,liu2025hybridvla,team2024octo,zhang2026ma}, or autoregressive action tokenization~\cite{pertsch2025fast,liu2025faster,bu2025univla,zhao2025cot,kim2024openvla}.
While these approaches substantially improve generalization and action modeling, failure recovery remains relatively underexplored.
CARE is orthogonal to the underlying VLA architecture and is instantiated on representative continuous and tokenized backbones, $\pi_0$ and $\pi_0$-FAST.

\subsection{Robotic Manipulation Benchmarks}
Existing benchmarks have advanced from short-horizon single-arm tasks~\cite{james2020rlbench,yu2020meta}, long-horizon multi-tasking and everyday environments~\cite{mees2022calvin,liu2023libero,mu2021maniskill,zhang2025vlabench,nasiriany2024robocasa,li2023behavior,shen2021igibson}, and real-world cross-embodiment generalization~\cite{o2024open} to multi-arm collaborative manipulation~\cite{chen2025robotwin,qin2025robofactory}. 
However, most benchmarks focus on task execution from predefined initial states and provide limited support for evaluating recovery from intermediate failures. FSR-Bench fills this gap by evaluating self-correction in dual-arm manipulation under execution deviations.

\begin{figure*}[t]
    \centering
    \includegraphics[width=\textwidth]{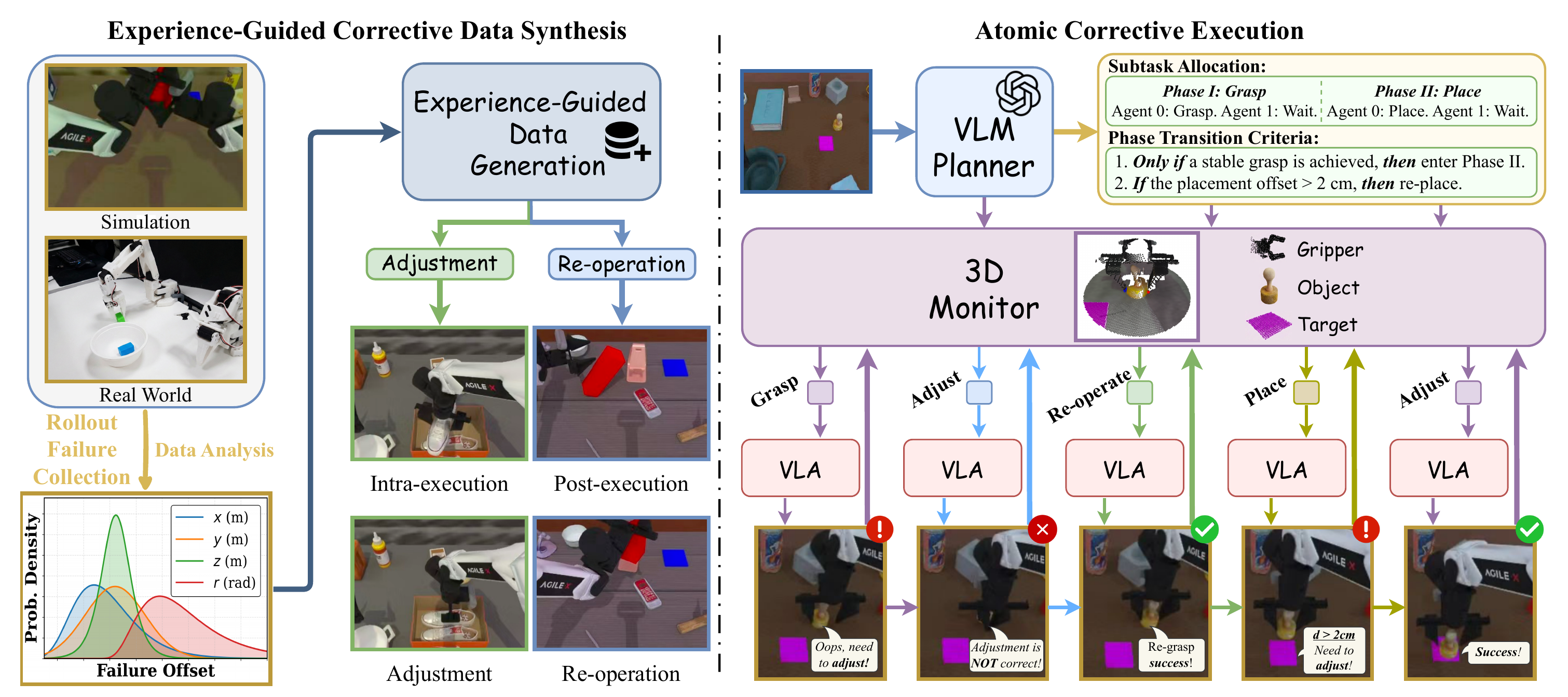}
    \caption{
    Overview of CARE.
    CARE first models stage-conditioned deviations from failed executions and uses them to generate corrective demonstrations.
    At inference time, stage-wise planning and 3D monitoring select nominal or corrective atomic instructions for a unified VLA executor.
    }
    \label{fig:method}
\end{figure*}

\section{Method}

\label{sec:method}

\subsection{Overview}

CARE addresses two complementary aspects of failure recovery:
\emph{where recovery supervision should come from} and
\emph{when corrective behavior should be invoked}.
First, Experience-Guided Corrective Data Synthesis models stage-conditioned deviations observed in failed executions and uses them to generate corrective demonstrations.
Second, Atomic Corrective Execution decomposes a task into atomic stages and uses 3D geometric monitoring to select nominal or corrective instructions during execution.
Both nominal and corrective instructions are executed by the same VLA policy.
Figure~\ref{fig:method} illustrates the overall framework.

\subsection{Experience-Guided Corrective Data Synthesis}
\label{sec:data_synthesis}

\subsubsection{Failure Modeling}

We begin by executing nominal atomic plans without corrective supervision and collect failure cases from 100 rollouts in simulation and the real world. A failure is recorded when an atomic stage does not satisfy its termination condition.

For a failure occurring at stage $k_i$, we characterize the deviation at the corresponding stage-critical event as
\begin{equation}
    d_i = (\Delta \tau_i, \Delta R_i),
\end{equation}
where $\Delta \tau_i \in \mathbb{R}^3$ and $\Delta R_i$ denote the relative translational and rotational deviations, respectively. We model rotational deviations as yaw offsets $\Delta r_i$, with $\Delta R_i = R_z(\Delta r_i)$, where $R_z(\cdot)$ denotes rotation about the vertical axis. The deviations are measured in relative coordinates: for grasping, they describe object--gripper misalignment at gripper closing; for placement, they describe object--target misalignment at gripper opening.

Because different stages exhibit different failure patterns, we model a stage-conditioned deviation distribution
\begin{equation}
    p(d \mid k).
\end{equation}

For each task and atomic stage, we independently fit each scalar component $(\Delta x,\Delta y,\Delta z,\Delta r)$ to Gaussian, Beta, Gamma, Weibull, Log-Normal, and Uniform candidate distributions, and select the best fit using AIC~\cite{akaike1974new} and the KS test~\cite{massey1951kolmogorov}.

\subsubsection{Corrective Data Synthesis}

To generate a new failure state at stage $k$, we sample
\begin{equation}
    d = (\Delta \tau, \Delta R) \sim p(d \mid k),
\end{equation}
and perturb the stage-relevant relative pose:
\begin{equation}
    \tau_k \leftarrow \tau_k + \Delta \tau,
    \qquad
    R_k \leftarrow \Delta R R_k.
\end{equation}

The system is then rolled forward under physical dynamics. Thus, synthesized failures are newly generated physical states rather than direct replays of previously collected failures. In simulation, an IK/motion-planning oracle generates corrective demonstrations from these states, whereas real-world demonstrations are collected through human teleoperation. Both consist of short corrective segments that terminate once the atomic recovery target is reached.

\begin{figure*}[t]
    \centering
    \includegraphics[width=\textwidth]{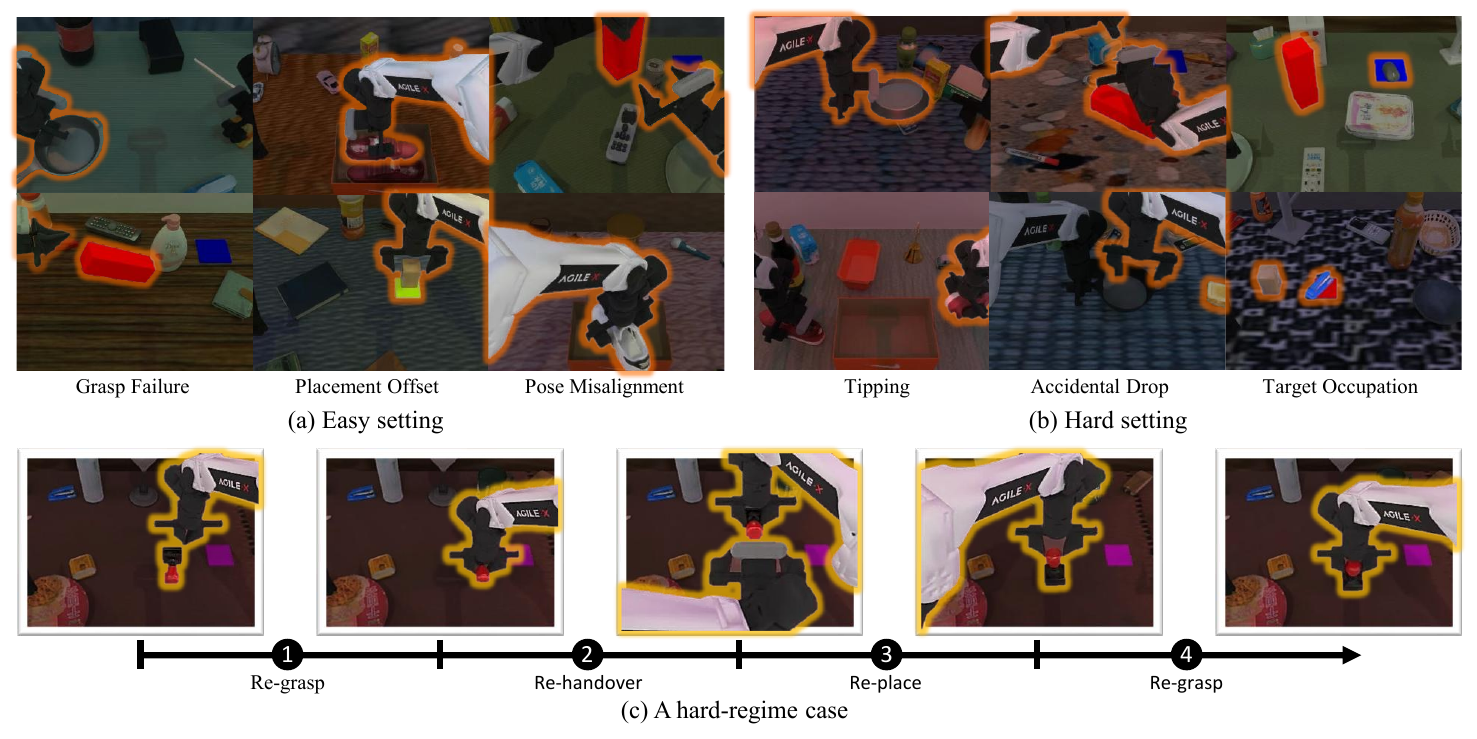}
    \caption{\textbf{FSR-Bench.}
    Recovery scenarios are grouped into two regimes by corrective complexity:
    (a) \textbf{Easy}, including local deviations that typically admit single-step correction, such as \textit{Grasp Failure}, \textit{Placement Offset}, and \textit{Pose Misalignment};
    and (b) \textbf{Hard}, including structural anomalies that require multi-step recovery or multi-arm coordination, such as \textit{Tipping}, \textit{Accidental Drop}, and \textit{Target Occupation}.
    Each column shows two representative initializations.
    (c) \textbf{A hard-regime case}, where repeated monitor--execute--verify transitions select a coordinated sequence of bimanual corrective skills online.}
    \label{fig:benchmark}
\end{figure*}

\subsection{Atomic Corrective Execution Framework}
\label{sec:atomic_execution}

Learning corrective behaviors alone does not determine when they should be invoked.
CARE therefore performs recovery at the atomic-skill level.
A VLM planner decomposes the task and specifies stage-level recovery semantics, while a 3D geometric monitor determines whether to continue the current stage, trigger adjustment, or trigger re-operation.
The planner and monitor only select the atomic instruction; all low-level actions, including corrective actions, are generated by the same VLA executor.

\subsubsection{VLM Planner}
We use GPT-4.1 (temperature 0) to decompose the
high-level instruction $\ell$ and initial global-view image $I^g_0$
into $K$ atomic stages.
For each stage $k$, it produces
\begin{equation}
\Pi_k = (u_k, \gamma_k, \psi_k, \mathcal{U}_k),
\end{equation}
where $u_k$ is the nominal atomic instruction, $\gamma_k$ specifies the geometric cues relevant to recovery, $\psi_k$ is the stage-termination predicate, and
\begin{equation}
\mathcal{U}_k =
\mathcal{U}_k^{\mathrm{adj}}
\cup
\mathcal{U}_k^{\mathrm{re}}
\end{equation}
contains instructions for intra-execution adjustment and post-execution re-operation.


The planner is queried once at task initialization. The resulting stage plans and candidate instructions remain fixed, while CARE updates the active stage \(k_t\) and selects the executed instruction online. This separates semantic task decomposition from high-frequency visuomotor control and avoids repeated VLM calls during recovery.

\subsubsection{3D Geometric Monitor}

The 3D monitor is a point-cloud predicate checker rather than a learned recovery policy or a VLM-based judge.
It continuously tracks gripper open/close states and invokes geometric
perception at stage-critical transitions, such as grasp closure or object
release, or shortly before a predicted transition when intra-execution
adjustment is enabled.

For each triggered check, SAM~3~\cite{carion2025sam} extracts masks for the gripper, manipulated object, and target region from the RGB observations, and Depth Anything~3~\cite{lin2025depth} provides depth estimates.
These observations are fused into the gripper, object, and target point clouds
$\mathcal{P}_{g,t}$, $\mathcal{P}_{o,t}$, and $\mathcal{P}_{\mathrm{tar},t}$.
The monitor then computes a compact geometric state
\begin{equation}
q_t =
\Phi(
\mathcal{P}_{g,t},
\mathcal{P}_{o,t},
\mathcal{P}_{\mathrm{tar},t}),
\end{equation}
with stage-relevant relations such as gripper--object alignment, object--target displacement, and post-contact stability.


The monitor uses skill-level geometric predicates shared across tasks composed of known atomic skills, with numerical tolerances calibrated from rollout statistics.
A new atomic skill requires defining its predicate template once, which can then be reused across tasks.

Conditioned on the active plan $\Pi_{k_t}$ and geometric state $q_t$, the monitor selects
\begin{equation}
\tilde{u}_t =
f_{\mathrm{mon}}(\Pi_{k_t},q_t),
\qquad
\tilde{u}_t
\in
\{u_{k_t}\}
\cup
\mathcal{U}_{k_t}^{\mathrm{adj}}
\cup
\mathcal{U}_{k_t}^{\mathrm{re}}.
\end{equation}


Consistent geometry retains \(u_{k_t}\); in-stage deviations trigger adjustment, while failed outcomes trigger re-operation. After each atomic execution, the monitor reapplies Eq.~(8): CARE advances to \(k_t+1\) when \(\psi_{k_t}\) holds; otherwise, it retains \(k_t\) and retries until the retry budget is exhausted, after which recovery fails. Repeated monitor--execute--verify transitions thereby generate multi-step recovery over the fixed plan without a predefined sequence.

\subsubsection{VLA Executor}

The VLA executor receives the current observation $o_t$ together with the selected atomic instruction:
\begin{equation}
a_t \sim \pi(a_t \mid o_t,\tilde{u}_t).
\end{equation}
Here, $o_t$ contains the robot state together with the global and wrist-view RGB observations.
The same VLA policy executes nominal, adjustment, and re-operation instructions; CARE therefore invokes learned corrective behaviors without switching to a separate recovery policy.

\subsection{FSR-Bench}
\label{sec:fsr_bench}

FSR-Bench evaluates failure recovery independently of nominal
task execution.
Instead of starting from clean initial states, each episode begins
from an intermediate failure state and requires the policy to
restore a task-feasible state.
The benchmark contains 36 recovery scenarios across five tasks,
organized into two regimes by corrective complexity.
\emph{Easy} contains 21 local failures that typically admit a
single corrective operation, such as grasp failure, placement
offset, and pose misalignment.
\emph{Hard} contains 15 structural failures that require multi-step
correction or coordinated multi-arm recovery, such as tipping,
accidental drop, and target occupation.
Representative scenarios are shown in Fig.~\ref{fig:benchmark}.

\paragraph{Training and Test Distributions}
To decouple recovery training from the benchmark test distribution, the recovery training data used in FSR-Bench are collected by uniformly sampling perturbations within the predefined perturbation range of each recovery scenario.

In contrast, benchmark evaluation reflects failures arising
during actual policy execution. We collect failed rollouts from
a mixture of nominal policies without recovery and estimate
a stage-conditioned empirical failure distribution
\begin{equation}
p_{\mathrm{fail}}(d \mid k),
\end{equation}
where $d$ denotes the geometric deviation at atomic stage $k$.

At evaluation time, we sample
\begin{equation}
d \sim p_{\mathrm{fail}}(d \mid k),
\end{equation}
inject the sampled deviation into the corresponding stage of a nominal execution, and roll the system forward under physical dynamics.
The resulting state, rather than the injected pose itself, is used as the recovery initialization.
Each evaluation episode contains a new failure state rather than a replay of a collected failure.
All baseline and CARE variants use the same uniformly sampled corrective training data and evaluation initializations.
Thus, their comparison evaluates atomic corrective execution under failures generated from a mixture of nominal policies.

Evaluation episodes additionally include variations in object geometry, table texture, lighting, distractor objects, and object placement.
For each scenario, all methods are evaluated on the same pre-generated failure initializations.

\paragraph{Evaluation Protocol}
Each method is evaluated over $N_{\mathrm{eval}}=100$
randomized trials per task--regime pair, with initializations
drawn from the corresponding recovery scenarios.
Recovery Success Rate (RSR) is defined as
\begin{equation}
\mathrm{RSR}
=
\frac{1}{N_{\mathrm{eval}}}
\sum_{j=1}^{N_{\mathrm{eval}}}
\mathbb{I}[r_j=1],
\end{equation}

where $r_j=1$ if the scenario-specific recovery target is reached
within its prescribed horizon. For each scenario, its target and
horizon are shared across methods.
For each task and regime, RSR is computed over 100 trials sampled
from the corresponding constituent failure states.

\section{Experiments}
\label{sec:experiment}
\definecolor{shading}{gray}{0.9}

\subsection{Experimental Setup}
\label{sec:exp_setup}

\paragraph{Tasks and baselines}
We evaluate CARE on seven hard bimanual tasks from RoboTwin~2.0~\cite{chen2025robotwin}, three multi-arm tasks from RoboFactory~\cite{qin2025robofactory}, five FSR-Bench recovery tasks, and four real-world dual-arm tasks.
For RoboTwin~2.0, we use the \texttt{demo\_randomized} setting for training and evaluation.
We instantiate CARE on $\pi_0$~\cite{black2024pi_0} and $\pi_0$-FAST~\cite{pertsch2025fast}, and compare against their vanilla counterparts, ACT~\cite{zhao2023learning}, RDT-1B~\cite{liu2024rdt}, DP~\cite{chi2025diffusion}, and DP3~\cite{ze20243d}.
On FSR-Bench, we additionally instantiate CARE on RDT-1B.

\paragraph{Training protocol}
For standard simulation tasks, each method uses 150 nominal expert demonstrations and 50 additional trajectories per error type.
CARE uses experience-guided corrective trajectories, while baselines use an equal number of nominal atomic-stage trajectories.
The corrective data are synthesized from stage-conditioned failure distributions estimated from 100 preliminary rollouts.
Policies are trained for 40{,}000 gradient steps with a batch size of 32 and an action horizon of 50 on two NVIDIA A800 SXM4 80GB GPUs.
For FSR-Bench, all variants use 50 uniformly sampled corrective trajectories per recovery type and are trained for 20{,}000 steps with the same batch size and horizon.

\paragraph{Real-world setup and evaluation}
We evaluate four bimanual tasks on a dual-arm LeRobot SO-101 platform with 12 DoF (6 per arm), one global RGB camera, and one wrist RGB camera per arm, all operating at 30\,Hz.
For each task, we collect 50 nominal demonstrations and 20 additional trajectories per error type, following the same matched-data protocol as in simulation.
We use 100 preliminary executions per task to estimate the empirical failure distribution.
Real-world task success is evaluated over 100 trials per task.

\begin{table}[h!]
\centering
\caption{Performance on bimanual tasks in RoboTwin~2.0 (Hard)~\cite{chen2025robotwin}.
Short/Long denote short-horizon ($<$200 steps) and long-horizon (200--400 steps).
LP: Lift Pot; SS: Stamp Seal; PBS: Place Bread Skillet;
PDS: Place Dual Shoes; POC: Put Object Cabinet;
HB: Handover Block; DBB: Dump Bin Bigbin.}
\label{tab:robotwin_hard_results}
\footnotesize
\setlength{\tabcolsep}{2.2pt}
\renewcommand{\arraystretch}{0.82}
\resizebox{1\columnwidth}{!}{%
\begin{tabular}{lcccccccc}
\toprule
\multirow{2}{*}{Model} & \multicolumn{3}{c}{Short} & \multicolumn{4}{c}{Long} & \multirow{2}{*}{Avg.} \\
\cmidrule(lr){2-4} \cmidrule(lr){5-8}
& LP & SS & PBS & PDS & POC & HB & DBB & \\
\midrule
ACT~\cite{zhao2023learning}
& 46\% & 0\% & 5\% & 0\% & 23\% & 0\% & 74\% & 21.1\% \\

RDT-1B~\cite{liu2024rdt}
& 15\% & 2\% & 2\% & 2\% & 23\% & 7\% & 75\% & 18.0\% \\

DP~\cite{chi2025diffusion}
& 52\% & 0\% & 6\% & 4\% & 42\% & 21\% & 53\% & 25.4\% \\

DP3~\cite{ze20243d}
& 27\% & 1\% & 9\% & 7\% & 52\% & 14\% & 87\% & 28.1\% \\
\midrule
$\pi_0$-FAST~\cite{pertsch2025fast}
& 18\% & 2\% & 2\% & 10\% & 23\% & \textbf{24\%} & 43\% & 17.4\% \\

\rowcolor{shading}
$\pi_0$-FAST (CARE)
& \textbf{40\%} & \textbf{17\%} & \textbf{28\%} & \textbf{33\%} & \textbf{32\%} & 16\% & \textbf{66\%} & \textbf{33.1\%} \\
\midrule
$\pi_0$~\cite{black2024pi_0}
& 57\% & 3\% & 28\% & 14\% & 59\% & 27\% & 89\% & 39.6\% \\

\rowcolor{shading}
$\pi_0$ (CARE)
& \textbf{91\%} & \textbf{29\%} & \textbf{41\%} & \textbf{40\%} & \textbf{68\%} & \textbf{62\%} & \textbf{94\%} & \textbf{60.7\%} \\
\bottomrule
\end{tabular}%
}
\end{table}

\begin{table}[!h]
    \centering
    \caption{Performance on multi-arm collaboration tasks in RoboFactory~\cite{qin2025robofactory}.}
    \label{tab:multi_robot_results}
    \footnotesize
    \setlength{\tabcolsep}{2.2pt}
    \renewcommand{\arraystretch}{0.95}
    \resizebox{1\columnwidth}{!}{%
    \begin{tabular}{lcccc}
        \toprule
        Model & \shortstack{Place food\\(Two Arms)} & \shortstack{Stack cubes three\\(Three Arms)} & \shortstack{Take photo\\(Four Arms)} & Avg. \\
        \midrule
        DP~\cite{chi2025diffusion} & 20\% & 22\% & 20\% & 20.7\% \\
        \midrule
        $\pi_0$-FAST~\cite{pertsch2025fast} & 27\% & 9\% & 44\% & 26.7\% \\
        \rowcolor{shading}
        $\pi_0$-FAST~(CARE) & \textbf{40\%} & \textbf{17\%} & \textbf{51\%} & \textbf{36.0\%} \\
        \midrule
        $\pi_0$~\cite{black2024pi_0} & 56\% & 48\% & 82\% & 62.0\% \\
        \rowcolor{shading}
        $\pi_0$~(CARE) & \textbf{73\%} & \textbf{61\%} & \textbf{88\%} & \textbf{74.0\%} \\
        \bottomrule
    \end{tabular}%
    }
    \vspace{-1mm}
\end{table}

\subsection{Main Results in Simulation}
\label{sec:exp_main_results}

Table~\ref{tab:robotwin_hard_results} evaluates CARE on hard bimanual
tasks in RoboTwin~2.0, where recovery requires precise geometric
correction under challenging interactions. CARE substantially improves
both VLA backbones, raising average success rates from 39.6\% to
60.7\% (+21.1 points) for $\pi_0$ and from 17.4\% to 33.1\%
(+15.7 points) for $\pi_0$-FAST. These gains span short- and
long-horizon tasks. Table~\ref{tab:multi_robot_results} further shows
CARE's benefits for multi-arm collaboration in RoboFactory. Across
two-, three-, and four-arm tasks, CARE improves both backbones, with
average absolute gains of +9.3 points for $\pi_0$-FAST and +12.0
points for $\pi_0$. These results demonstrate CARE's
failure-recovery capability across task horizons and multi-arm settings.

\begin{table*}[t]
\centering
\caption{Performance comparison on FSR-Bench. DSR: Dual Shoes Recovery; SSR: Seal Stamp Recovery; SPR: Skillet Placement Recovery; BHR: Block Handover Recovery; PLR: Pot Lift Recovery.}
\label{tab:vla_results}
\footnotesize
\setlength{\tabcolsep}{2pt}
\renewcommand{\arraystretch}{0.85}
\begin{tabular*}{\textwidth}{@{\extracolsep{\fill}}lcccccccccccc}
\toprule
\multirow{2}{*}{Task} 
& \multicolumn{2}{c}{RDT-1B~\cite{liu2024rdt}}
& \multicolumn{2}{c}{RDT-1B (CARE)}
& \multicolumn{2}{c}{$\pi_0$-FAST~\cite{pertsch2025fast}}
& \multicolumn{2}{c}{$\pi_0$-FAST (CARE)}
& \multicolumn{2}{c}{$\pi_0$~\cite{black2024pi_0}}
& \multicolumn{2}{c}{$\pi_0$ (CARE)} \\
\cmidrule(lr){2-3} \cmidrule(lr){4-5}
\cmidrule(lr){6-7} \cmidrule(lr){8-9}
\cmidrule(lr){10-11} \cmidrule(lr){12-13}
& Easy & Hard & Easy & Hard
& Easy & Hard & Easy & Hard
& Easy & Hard & Easy & Hard \\
\midrule
DSR
& 42\% & 2\% & \textbf{51\%} & \textbf{4\%}
& 44\% & 2\% & \textbf{48\%} & \textbf{11\%}
& 63\% & 7\% & \textbf{69\%} & \textbf{9\%} \\

SSR
& 10\% & 2\% & \textbf{11\%} & \textbf{2\%}
& 27\% & 5\% & \textbf{30\%} & \textbf{9\%}
& 36\% & 18\% & \textbf{54\%} & \textbf{24\%} \\

SPR
& 11\% & 2\% & \textbf{13\%} & \textbf{14\%}
& 12\% & 6\% & \textbf{16\%} & \textbf{12\%}
& 25\% & 32\% & \textbf{31\%} & \textbf{40\%} \\

BHR
& 36\% & 0\% & \textbf{39\%} & \textbf{2\%}
& 35\% & 2\% & \textbf{45\%} & \textbf{4\%}
& 40\% & 11\% & \textbf{54\%} & \textbf{18\%} \\

PLR
& 10\% & 7\% & \textbf{12\%} & \textbf{8\%}
& 4\% & 8\% & \textbf{24\%} & \textbf{18\%}
& 30\% & 11\% & \textbf{78\%} & \textbf{15\%} \\
\midrule
Avg.
& 21.8\% & 2.6\% & \textbf{25.2\%} & \textbf{6.0\%}
& 24.4\% & 4.6\% & \textbf{32.6\%} & \textbf{10.8\%}
& 38.8\% & 15.8\% & \textbf{57.2\%} & \textbf{21.2\%} \\
\bottomrule
\vspace{-2mm}
\end{tabular*}
\end{table*}

\begin{figure*}[!t] 
    \centering
    \includegraphics[width=1.0\linewidth]{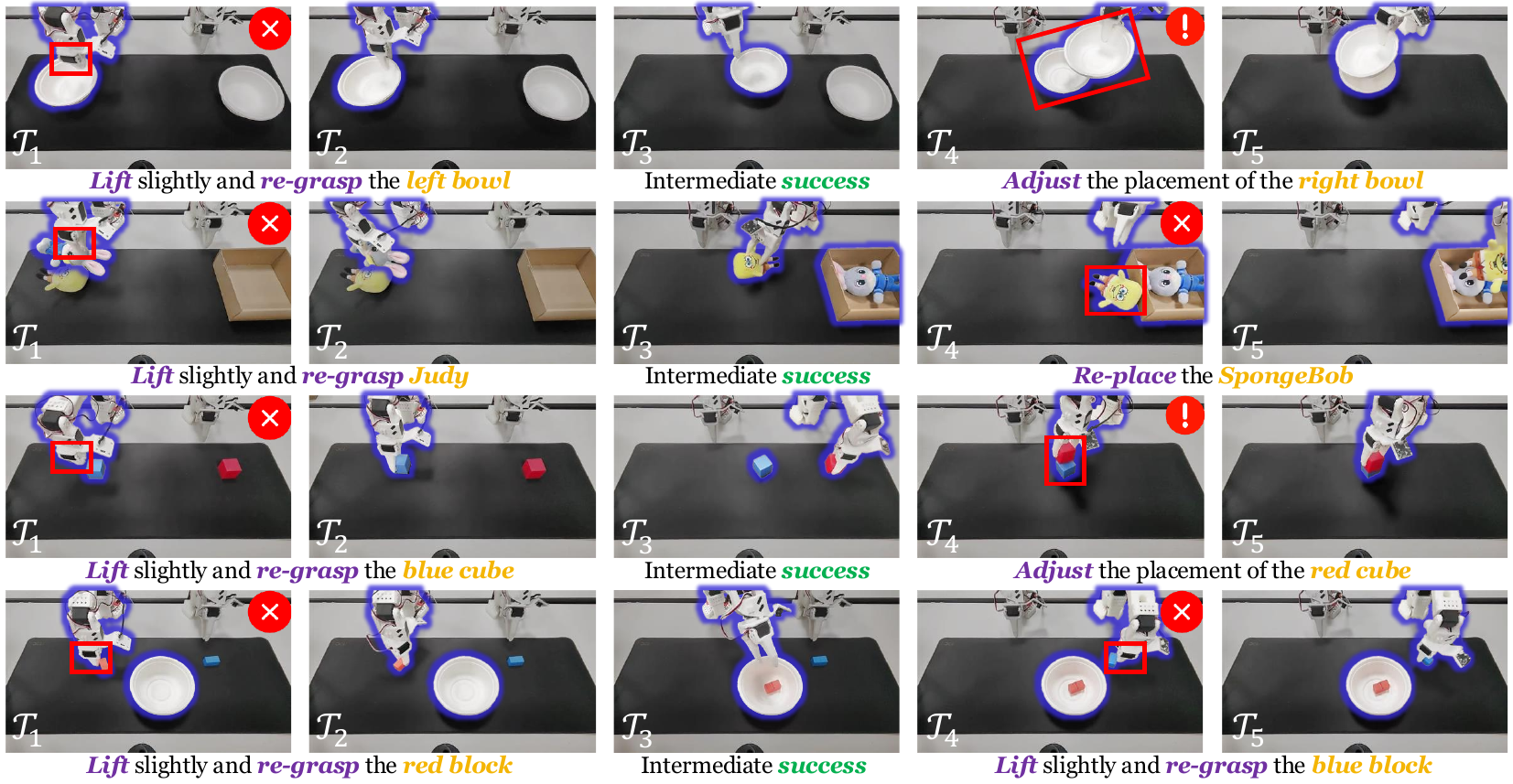}
    \caption{Qualitative results on four real-world dual-arm SO-101 tasks. From top to bottom: Stack Two Bowls, Pass Two Toys, Stack Two Cubes, and Place Two Cubes. \raisebox{-0.1em}{\includegraphics[height=1em]{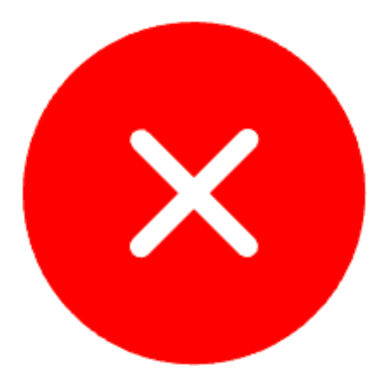}}\textbf{: Requiring post-execution re-operation; }\raisebox{-0.1em}{\includegraphics[height=1em]{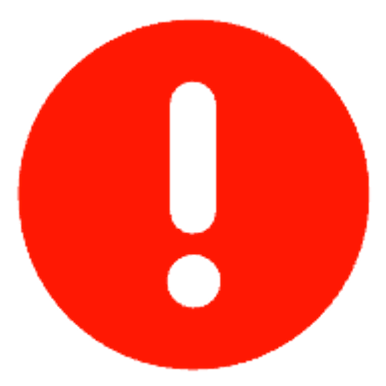}}\textbf{: Requiring intra-execution adjustment.}}
    \label{fig:real_world}
    \vspace{-1mm}
    
\end{figure*}

\subsection{Results in FSR-Bench}

Table~\ref{tab:vla_results} shows that CARE's Atomic Corrective Execution consistently improves recovery across all three VLA backbones and both regimes. To isolate execution, experience-guided synthesis is disabled for this benchmark, and each vanilla/CARE pair uses the same 50 uniformly sampled corrective atomic trajectories per recovery type. On average, RDT-1B improves from 21.8\% to 25.2\% in the Easy regime and from 2.6\% to 6.0\% in the Hard regime. $\pi_0$-FAST improves from 24.4\% to 32.6\% in the Easy regime and from 4.6\% to 10.8\% in the Hard regime. The largest gains are achieved by $\pi_0$, whose average RSR rises from 38.8\% to 57.2\% in the Easy regime and from 15.8\% to 21.2\% in the Hard regime. The consistent gains across backbones suggest that the execution mechanism is not model-specific. Overall, matched corrective supervision alone is insufficient for robust recovery, which also requires atomic corrective execution.

\subsection{Main Results in Real World}

We validate CARE on a real-world SO-101 dual-arm robot
across four long-horizon manipulation tasks
(Fig.~\ref{fig:real_world}).
For system-level comparison against a dedicated recovery method,
we reproduce the complete FailSafe~\cite{lin2025failsafe} pipeline
on the same platform with identical VLA backbones and evaluation protocol.
Table~\ref{tab:real_world_so101} reports success rates over 100 trials
per task.
CARE improves the average success rate of $\pi_0$ from 30.8\% to
50.0\% and that of $\pi_0$-FAST from 18.8\% to 31.3\%, outperforming
the dedicated FailSafe recovery baseline by 13.7 and 7.3 points,
respectively.

\begin{table}[t]
    \centering
    \caption{Success rates on SO-101 dual-arm robot tasks.}
    \label{tab:real_world_so101}
    \footnotesize
    \setlength{\tabcolsep}{1.0pt}
    \renewcommand{\arraystretch}{0.72}
    \begin{tabular*}{\columnwidth}{@{\extracolsep{\fill}}lcccccc@{}}
        \toprule
        \multirow{2}{*}{Task}
        & \multicolumn{3}{c}{$\pi_0$~\cite{black2024pi_0}}
        & \multicolumn{3}{c}{$\pi_0$-FAST~\cite{pertsch2025fast}} \\
        \cmidrule(lr){2-4}
        \cmidrule(lr){5-7}
        & Base & FailSafe & CARE
        & Base & FailSafe & CARE \\
        \midrule
        Stack Two Cubes
        & 2\%  & 6\%  & \textbf{14\%}
        & 0\%  & 2\%  & \textbf{6\%}  \\
        Stack Two Bowls
        & 44\% & 48\% & \textbf{62\%}
        & 27\% & 36\% & \textbf{46\%} \\
        Pass Two Toys
        & 15\% & 21\% & \textbf{42\%}
        & 7\%  & 11\% & \textbf{17\%} \\
        Place Two Cubes
        & 62\% & 70\% & \textbf{82\%}
        & 41\% & 47\% & \textbf{56\%} \\
        \midrule
        Avg.
        & 30.8\% & 36.3\% & \textbf{50.0\%}
        & 18.8\% & 24.0\% & \textbf{31.3\%} \\
        \bottomrule
    \end{tabular*}
    \vspace{-2mm}
\end{table}

\section{Ablation Study}
\label{sec:ablation}
We conduct ablation studies on selected simulation tasks from RoboTwin 2.0~\cite{chen2025robotwin} and RoboFactory~\cite{qin2025robofactory}, as well as real-world tasks on the dual-arm SO-101 platform, following the corresponding experimental configurations in Sec.~\ref{sec:exp_setup}. In Sec.~\ref{Com}, for each error type, the \textbf{Data} setting replaces the 50 additional nominal atomic-stage trajectories with 50 corrective trajectories, keeping the amount of added training data fixed. In Sec.~\ref{Col}, both the \textbf{Uniform Random} and \textbf{Experience-Guided} strategies use 50 corrective trajectories per error type under identical training settings.

\begin{table}[t]
    \centering
    \caption{Ablation of CARE components. PF: Place Food; PDS: Place Dual Shoes; LP: Lift Pot; PBS: Place Bread Skillet; SS: Stamp Seal.}
    \label{tab:ablation_combined}
    \setlength{\aboverulesep}{0pt}
    \setlength{\belowrulesep}{0pt}
    \setlength{\tabcolsep}{2pt}
    \scriptsize
    \resizebox{1\columnwidth}{!}{%
    \begin{tabular}{lcccccccc}
        \toprule
        \multirow{2}{*}{Model} & \multicolumn{2}{c}{Method} & \multicolumn{6}{c}{Task} \\
        \cmidrule(lr){2-3} \cmidrule(lr){4-9}
        & Execution & Data & PF & PDS & LP & PBS & SS & Avg. \\
        \midrule

        \multirow{4}{*}{$\pi_0$~\cite{black2024pi_0}}
          & -- & -- & 56\% & 14\% & 57\% & 28\% & 3\% & 31.6\% \\
          & \checkmark & -- & 58\% & 22\% & 58\% & 30\% & 11\% & 35.8\% \\
          & -- & \checkmark & 61\% & 33\% & 62\% & 26\% & 22\% & 40.8\% \\
          & \checkmark & \checkmark & \textbf{73\%} & \textbf{40\%} & \textbf{91\%} & \textbf{41\%} & \textbf{29\%} & \textbf{54.8\%} \\
        \midrule

        \multirow{4}{*}{$\pi_0$-FAST~\cite{pertsch2025fast}}
          & -- & -- & 27\% & 10\% & 18\% & 2\% & 2\% & 11.8\% \\
          & \checkmark & -- & 30\% & 11\% & 18\% & 6\% & 4\% & 13.8\% \\
          & -- & \checkmark & 32\% & 31\% & 23\% & 6\% & 12\% & 20.8\% \\
          & \checkmark & \checkmark & \textbf{40\%} & \textbf{33\%} & \textbf{40\%} & \textbf{28\%} & \textbf{17\%} & \textbf{31.6\%} \\
        \bottomrule
    \end{tabular}%
    }
    \vspace{-2mm}
\end{table}

\subsection{Ablation of Components}
\label{Com}
Table~\ref{tab:ablation_combined} ablates CARE's two core components: corrective training data (\textbf{Data}) and the atomic corrective execution framework (\textbf{Execution}). For both backbones, \textbf{Data} provides the larger gain. On $\pi_0$, \textbf{Data} alone improves the average success rate from 31.6\% to 40.8\%~(+9.2 points), while \textbf{Execution} alone raises it to 35.8\%~(+4.2 points). A similar trend holds for $\pi_0$-FAST, where \textbf{Data} alone improves performance from 11.8\% to 20.8\%~(+9.0 points), whereas \textbf{Execution} alone raises it to 13.8\%~(+2.0 points). Full CARE achieves the best results, outperforming the respective baselines by 23.2 points on $\pi_0$ and 19.8 points on $\pi_0$-FAST. Because full CARE and the data-only variant use identical nominal and corrective training data, their 14.0- and 10.8-point gaps isolate the additional benefit of online atomic corrective execution beyond corrective supervision alone.


\subsection{Ablation of Corrective Data Collection}
\label{Col}
\begin{table}[h!]
    \centering
    \caption{Ablation of corrective data collection. SCT: Stack Cubes Three; PDS: Place Dual Shoes; LP: Lift Pot; SS: Stamp Seal; HB: Handover Block.}
    \label{tab:sampling_comparison}
    \footnotesize
    \setlength{\tabcolsep}{1.6pt}
    \renewcommand{\arraystretch}{0.72}
    \resizebox{1\columnwidth}{!}{%
    \begin{tabular}{lcccc}
        \toprule
        \multirow{2}{*}{Task} & \multicolumn{2}{c}{Uniform Random} & \multicolumn{2}{c}{Experience-Guided (Ours)} \\
        \cmidrule(lr){2-3} \cmidrule(lr){4-5}
        & $\pi_0$~(CARE) & $\pi_0$-FAST~(CARE) & $\pi_0$~(CARE) & $\pi_0$-FAST~(CARE) \\
        \midrule
        SCT & 53\% & 5\%  & \textbf{61\%} & \textbf{17\%}  \\
        PDS & 27\% & 24\% & \textbf{40\%} & \textbf{33\%}  \\
        LP  & 66\% & 27\% & \textbf{91\%} & \textbf{40\%}  \\
        SS  & 20\% & 7\%  & \textbf{29\%} & \textbf{17\%} \\
        HB  & 56\% & 16\% & \textbf{62\%} & 16\% \\
        \midrule
        Avg. & 44.4\% & 15.8\% & \textbf{56.6\%} {\color{blue}(+12.2 pts)} & \textbf{24.6\%} {\color{blue}(+8.8 pts)} \\
        \bottomrule
    \end{tabular}%
    }
    \vspace{-1mm}
\end{table}


To isolate the effect of the sampling distribution, we compare our Experience-Guided Corrective Data Synthesis with a FailSafe-style Uniform Random strategy~\cite{lin2025failsafe}. Both use identical deviation dimensions and ranges, numbers of corrective trajectories, and training settings; only the sampling distribution differs: uniform sampling versus the fitted $p(d\mid k)$.
As shown in Table~\ref{tab:sampling_comparison}, experience-guided sampling yields consistently higher average performance than uniform sampling, improving the average success rate by 12.2 points for $\pi_0$~(CARE) and 8.8 points for $\pi_0$-FAST~(CARE). These results suggest that uniformly sampled perturbations often miss physically meaningful long-tail failure modes, whereas empirically grounded synthesis provides more effective recovery supervision for closed-loop correction.

\subsection{Ablation of Monitoring Mechanisms}
\begin{table}[t]
    \centering
    \caption{Ablation of monitoring mechanisms on $\pi_0$. Tasks marked with $\dagger$ are real-world.}
    \label{tab:monitoring_ablation}
    \footnotesize
    \setlength{\tabcolsep}{1.5pt}
    \renewcommand{\arraystretch}{0.85}
    \resizebox{\columnwidth}{!}{%
    \begin{tabular}{lcccc}
        \toprule
        \multirow{2}{*}{Task}
        & \multicolumn{2}{c}{Monitoring Accuracy $\uparrow$}
        & \multicolumn{2}{c}{Latency (s/call) $\downarrow$} \\
        \cmidrule(lr){2-3} \cmidrule(lr){4-5}
        & VLM & 3D Monitor~(Ours)
        & VLM & 3D Monitor~(Ours) \\
        \midrule

        Lift Pot~(LP)
        & 80\% & \textbf{98\%}
        & 2.32 & \textbf{0.24} \\

        Place Dual Shoes~(PDS)
        & 85\% & \textbf{95\%}
        & 2.51 & \textbf{0.25} \\

        Handover Block~(HB)
        & 50\% & \textbf{96\%}
        & 4.53 & \textbf{0.24} \\

        Pass Two Toys~(PTT)$^\dagger$
        & 70\% & \textbf{86\%}
        & 3.77 & \textbf{0.24} \\

        Stack Two Bowls~(STB)$^\dagger$
        & 72\% & \textbf{94\%}
        & 3.89 & \textbf{0.24} \\

        \midrule
        Avg.
        & 71.4\%
        & \textbf{93.8\%}{\color{blue}~(+22.4 pts)}
        & 3.40
        & \textbf{0.24}{\color{blue}~($14.1\times$ faster)} \\
        \bottomrule
    \end{tabular}%
    }
    \vspace{-1.5mm}
\end{table}

We replace our 3D monitor with a GPT-4.1-based VLM monitor and evaluate both on the same triggered observations with manual decision labels. As shown in Table~\ref{tab:monitoring_ablation}, across three simulated and two real-world tasks, our monitor improves average accuracy from 71.4\% to 93.8\% (+22.4 points), while reducing end-to-end monitor latency, including perception, point-cloud construction, and predicate evaluation, from 3.40 s to 0.24 s ($14.1\times$ faster). These results demonstrate more accurate and efficient monitoring across simulation and real-world deployment.

\section{Conclusion}
We present CARE, a framework for improving failure recovery in
long-horizon multi-arm VLA manipulation. CARE treats execution
failures as structured supervision, combining Experience-Guided
Corrective Data Synthesis with 3D-monitor-driven atomic corrective
execution. We introduce FSR-Bench to evaluate recovery from
empirically grounded intermediate failure states. Across simulation,
recovery-specific, and real-world evaluations, CARE consistently
improves recovery and robustness over diverse VLA backbones.
These results suggest that robust embodied intelligence requires
both strong nominal execution and effective grounded recovery from
failures. Future work will integrate reinforcement learning into CARE
to further enhance recovery.


\bibliographystyle{IEEEtran}
\bibliography{references}

@String{Computer = "{IEEE} Computer" }

@String{Chelsea = "Chelsea" }

@article{xu2025can,
  title={Can we detect failures without failure data? uncertainty-aware runtime failure detection for imitation learning policies},
  author={Xu, Chen and Nguyen, Tony Khuong and Dixon, Emma and Rodriguez, Christopher and Miller, Patrick and Lee, Robert and Shah, Paarth and Ambrus, Rares and Nishimura, Haruki and Itkina, Masha},
  journal={arXiv preprint arXiv:2503.08558},
  year={2025}
}

@article{bu2024closed,
  title={Closed-loop visuomotor control with generative expectation for robotic manipulation},
  author={Bu, Qingwen and Zeng, Jia and Chen, Li and Yang, Yanchao and Zhou, Guyue and Yan, Junchi and Luo, Ping and Cui, Heming and Ma, Yi and Li, Hongyang},
  journal={Advances in Neural Information Processing Systems},
  volume={37},
  pages={139002--139029},
  year={2024}
}

@inproceedings{li2018slip,
  title={Slip detection with combined tactile and visual information},
  author={Li, Jianhua and Dong, Siyuan and Adelson, Edward},
  booktitle={2018 IEEE International Conference on Robotics and Automation (ICRA)},
  pages={7772--7777},
  year={2018},
  organization={IEEE}
}

@article{shi2024yell,
  title={Yell at your robot: Improving on-the-fly from language corrections},
  author={Shi, Lucy Xiaoyang and Hu, Zheyuan and Zhao, Tony Z and Sharma, Archit and Pertsch, Karl and Luo, Jianlan and Levine, Sergey and Finn, Chelsea},
  journal={arXiv preprint arXiv:2403.12910},
  year={2024}
}

@inproceedings{xia2025phoenix,
  title={Phoenix: A motion-based self-reflection framework for fine-grained robotic action correction},
  author={Xia, Wenke and Feng, Ruoxuan and Wang, Dong and Hu, Di},
  booktitle={Proceedings of the IEEE/CVF Conference on Computer Vision and Pattern Recognition},
  pages={6981--6990},
  year={2025}
}

@article{duan2024aha,
  title={Aha: A vision-language-model for detecting and reasoning over failures in robotic manipulation},
  author={Duan, Jiafei and Pumacay, Wilbert and Kumar, Nishanth and Wang, Yi Ru and Tian, Shulin and Yuan, Wentao and Krishna, Ranjay and Fox, Dieter and Mandlekar, Ajay and Guo, Yijie},
  journal={arXiv preprint arXiv:2410.00371},
  year={2024}
}

@inproceedings{dai2025racer,
  title={Racer: Rich language-guided failure recovery policies for imitation learning},
  author={Dai, Yinpei and Lee, Jayjun and Fazeli, Nima and Chai, Joyce},
  booktitle={2025 IEEE International Conference on Robotics and Automation (ICRA)},
  pages={15657--15664},
  year={2025},
  organization={IEEE}
}

@article{qi2026self,
  title={Self-Refining Vision Language Model for Robotic Failure Detection and Reasoning},
  author={Qi, Carl and Wang, Xiaojie and Yong, Silong and Sheng, Stephen and Mao, Huitan and Srinivasan, Sriram and Nambi, Manikantan and Zhang, Amy and Dattatreya, Yesh},
  journal={arXiv preprint arXiv:2602.12405},
  year={2026}
}

@article{black2024pi_0,
  title={pi\_0: A Vision-Language-Action Flow Model for General Robot Control},
  author={Black, Kevin and Brown, Noah and Driess, Danny and Esmail, Adnan and Equi, Michael and Finn, Chelsea and Fusai, Niccolo and Groom, Lachy and Hausman, Karol and Ichter, Brian and others},
  journal={arXiv preprint arXiv:2410.24164},
  year={2024}
}

@article{liu2024rdt,
  title={Rdt-1b: a diffusion foundation model for bimanual manipulation},
  author={Liu, Songming and Wu, Lingxuan and Li, Bangguo and Tan, Hengkai and Chen, Huayu and Wang, Zhengyi and Xu, Ke and Su, Hang and Zhu, Jun},
  journal={arXiv preprint arXiv:2410.07864},
  year={2024}
}

@article{wen2025dexvla,
  title={Dexvla: Vision-language model with plug-in diffusion expert for general robot control},
  author={Wen, Junjie and Zhu, Yichen and Li, Jinming and Tang, Zhibin and Shen, Chaomin and Feng, Feifei},
  journal={arXiv preprint arXiv:2502.05855},
  year={2025}
}

@article{liu2025hybridvla,
  title={Hybridvla: Collaborative diffusion and autoregression in a unified vision-language-action model},
  author={Liu, Jiaming and Chen, Hao and An, Pengju and Liu, Zhuoyang and Zhang, Renrui and Gu, Chenyang and Li, Xiaoqi and Guo, Ziyu and Chen, Sixiang and Liu, Mengzhen and others},
  journal={arXiv preprint arXiv:2503.10631},
  year={2025}
}

@article{team2024octo,
  title={Octo: An open-source generalist robot policy},
  author={Team, Octo Model and Ghosh, Dibya and Walke, Homer and Pertsch, Karl and Black, Kevin and Mees, Oier and Dasari, Sudeep and Hejna, Joey and Kreiman, Tobias and Xu, Charles and others},
  journal={arXiv preprint arXiv:2405.12213},
  year={2024}
}

@article{pertsch2025fast,
  title={Fast: Efficient action tokenization for vision-language-action models},
  author={Pertsch, Karl and Stachowicz, Kyle and Ichter, Brian and Driess, Danny and Nair, Suraj and Vuong, Quan and Mees, Oier and Finn, Chelsea and Levine, Sergey},
  journal={arXiv preprint arXiv:2501.09747},
  year={2025}
}

@article{liu2025faster,
  title={FASTer: Toward Efficient Autoregressive Vision Language Action Modeling via Neural Action Tokenization},
  author={Liu, Yicheng and Zhang, Shiduo and Dong, Zibin and Ye, Baijun and Yuan, Tianyuan and Yu, Xiaopeng and Yin, Linqi and Lu, Chenhao and Shi, Junhao and Yu, Luca Jiang-Tao and others},
  journal={arXiv preprint arXiv:2512.04952},
  year={2025}
}

@article{bu2025univla,
  title={Univla: Learning to act anywhere with task-centric latent actions},
  author={Bu, Qingwen and Yang, Yanting and Cai, Jisong and Gao, Shenyuan and Ren, Guanghui and Yao, Maoqing and Luo, Ping and Li, Hongyang},
  journal={arXiv preprint arXiv:2505.06111},
  year={2025}
}

@inproceedings{zhao2025cot,
  title={Cot-vla: Visual chain-of-thought reasoning for vision-language-action models},
  author={Zhao, Qingqing and Lu, Yao and Kim, Moo Jin and Fu, Zipeng and Zhang, Zhuoyang and Wu, Yecheng and Li, Zhaoshuo and Ma, Qianli and Han, Song and Finn, Chelsea and others},
  booktitle={Proceedings of the Computer Vision and Pattern Recognition Conference},
  pages={1702--1713},
  year={2025}
}

@article{kim2024openvla,
  title={Openvla: An open-source vision-language-action model},
  author={Kim, Moo Jin and Pertsch, Karl and Karamcheti, Siddharth and Xiao, Ted and Balakrishna, Ashwin and Nair, Suraj and Rafailov, Rafael and Foster, Ethan and Lam, Grace and Sanketi, Pannag and others},
  journal={arXiv preprint arXiv:2406.09246},
  year={2024}
}

@article{james2020rlbench,
  title={Rlbench: The robot learning benchmark \& learning environment},
  author={James, Stephen and Ma, Zicong and Arrojo, David Rovick and Davison, Andrew J},
  journal={IEEE Robotics and Automation Letters},
  volume={5},
  number={2},
  pages={3019--3026},
  year={2020},
  publisher={IEEE}
}

@inproceedings{yu2020meta,
  title={Meta-world: A benchmark and evaluation for multi-task and meta reinforcement learning},
  author={Yu, Tianhe and Quillen, Deirdre and He, Zhanpeng and Julian, Ryan and Hausman, Karol and Finn, Chelsea and Levine, Sergey},
  booktitle={Conference on robot learning},
  pages={1094--1100},
  year={2020},
  organization={PMLR}
}

@article{mees2022calvin,
  title={Calvin: A benchmark for language-conditioned policy learning for long-horizon robot manipulation tasks},
  author={Mees, Oier and Hermann, Lukas and Rosete-Beas, Erick and Burgard, Wolfram},
  journal={IEEE Robotics and Automation Letters},
  volume={7},
  number={3},
  pages={7327--7334},
  year={2022},
  publisher={IEEE}
}

@article{liu2023libero,
  title={Libero: Benchmarking knowledge transfer for lifelong robot learning},
  author={Liu, Bo and Zhu, Yifeng and Gao, Chongkai and Feng, Yihao and Liu, Qiang and Zhu, Yuke and Stone, Peter},
  journal={Advances in Neural Information Processing Systems},
  volume={36},
  pages={44776--44791},
  year={2023}
}

@article{mu2021maniskill,
  title={Maniskill: Generalizable manipulation skill benchmark with large-scale demonstrations},
  author={Mu, Tongzhou and Ling, Zhan and Xiang, Fanbo and Yang, Derek and Li, Xuanlin and Tao, Stone and Huang, Zhiao and Jia, Zhiwei and Su, Hao},
  journal={arXiv preprint arXiv:2107.14483},
  year={2021}
}

@article{nasiriany2024robocasa,
  title={Robocasa: Large-scale simulation of everyday tasks for generalist robots},
  author={Nasiriany, Soroush and Maddukuri, Abhiram and Zhang, Lance and Parikh, Adeet and Lo, Aaron and Joshi, Abhishek and Mandlekar, Ajay and Zhu, Yuke},
  journal={arXiv preprint arXiv:2406.02523},
  year={2024}
}

@inproceedings{li2023behavior,
  title={Behavior-1k: A benchmark for embodied ai with 1,000 everyday activities and realistic simulation},
  author={Li, Chengshu and Zhang, Ruohan and Wong, Josiah and Gokmen, Cem and Srivastava, Sanjana and Mart{\'\i}n-Mart{\'\i}n, Roberto and Wang, Chen and Levine, Gabrael and Lingelbach, Michael and Sun, Jiankai and others},
  booktitle={Conference on Robot Learning},
  pages={80--93},
  year={2023},
  organization={PMLR}
}

@inproceedings{shen2021igibson,
  title={iGibson 1.0: A simulation environment for interactive tasks in large realistic scenes},
  author={Shen, Bokui and Xia, Fei and Li, Chengshu and Mart{\'\i}n-Mart{\'\i}n, Roberto and Fan, Linxi and Wang, Guanzhi and P{\'e}rez-D’Arpino, Claudia and Buch, Shyamal and Srivastava, Sanjana and Tchapmi, Lyne and others},
  booktitle={2021 IEEE/RSJ International Conference on Intelligent Robots and Systems (IROS)},
  pages={7520--7527},
  year={2021},
  organization={IEEE}
}

@inproceedings{zhang2025vlabench,
  title={Vlabench: A large-scale benchmark for language-conditioned robotics manipulation with long-horizon reasoning tasks},
  author={Zhang, Shiduo and Xu, Zhe and Liu, Peiju and Yu, Xiaopeng and Li, Yuan and Gao, Qinghui and Fei, Zhaoye and Yin, Zhangyue and Wu, Zuxuan and Jiang, Yu-Gang and others},
  booktitle={Proceedings of the IEEE/CVF International Conference on Computer Vision},
  pages={11142--11152},
  year={2025}
}

@inproceedings{o2024open,
  title={Open x-embodiment: Robotic learning datasets and rt-x models: Open x-embodiment collaboration 0},
  author={O’Neill, Abby and Rehman, Abdul and Maddukuri, Abhiram and Gupta, Abhishek and Padalkar, Abhishek and Lee, Abraham and Pooley, Acorn and Gupta, Agrim and Mandlekar, Ajay and Jain, Ajinkya and others},
  booktitle={2024 IEEE International Conference on Robotics and Automation (ICRA)},
  pages={6892--6903},
  year={2024},
  organization={IEEE}
}

@article{chen2025robotwin,
  title={Robotwin 2.0: A scalable data generator and benchmark with strong domain randomization for robust bimanual robotic manipulation},
  author={Chen, Tianxing and Chen, Zanxin and Chen, Baijun and Cai, Zijian and Liu, Yibin and Li, Zixuan and Liang, Qiwei and Lin, Xianliang and Ge, Yiheng and Gu, Zhenyu and others},
  journal={arXiv preprint arXiv:2506.18088},
  year={2025}
}

@inproceedings{qin2025robofactory,
  title={Robofactory: Exploring embodied agent collaboration with compositional constraints},
  author={Qin, Yiran and Kang, Li and Song, Xiufeng and Yin, Zhenfei and Liu, Xiaohong and Liu, Xihui and Zhang, Ruimao and Bai, Lei},
  booktitle={Proceedings of the IEEE/CVF International Conference on Computer Vision},
  pages={10075--10085},
  year={2025}
}

@article{zhao2023learning,
  title={Learning fine-grained bimanual manipulation with low-cost hardware},
  author={Zhao, Tony Z and Kumar, Vikash and Levine, Sergey and Finn, Chelsea},
  journal={arXiv preprint arXiv:2304.13705},
  year={2023}
}

@article{chi2025diffusion,
  title={Diffusion policy: Visuomotor policy learning via action diffusion},
  author={Chi, Cheng and Xu, Zhenjia and Feng, Siyuan and Cousineau, Eric and Du, Yilun and Burchfiel, Benjamin and Tedrake, Russ and Song, Shuran},
  journal={The International Journal of Robotics Research},
  volume={44},
  number={10-11},
  pages={1684--1704},
  year={2025},
  publisher={Sage Publications Sage UK: London, England}
}

@article{ze20243d,
  title={3d diffusion policy: Generalizable visuomotor policy learning via simple 3d representations},
  author={Ze, Yanjie and Zhang, Gu and Zhang, Kangning and Hu, Chenyuan and Wang, Muhan and Xu, Huazhe},
  journal={arXiv preprint arXiv:2403.03954},
  year={2024}
}

@inproceedings{pan2025omnimanip,
  title={Omnimanip: Towards general robotic manipulation via object-centric interaction primitives as spatial constraints},
  author={Pan, Mingjie and Zhang, Jiyao and Wu, Tianshu and Zhao, Yinghao and Gao, Wenlong and Dong, Hao},
  booktitle={Proceedings of the Computer Vision and Pattern Recognition Conference},
  pages={17359--17369},
  year={2025}
}

@inproceedings{pan2025self,
  title={Self-correcting robot manipulation via Gaussian-splatted foresight},
  author={Pan, Shaohui and Xu, Yong and Xu, Ruotao and Zhou, Zihan and Wu, Si and Yu, Zhuliang},
  booktitle={Proceedings of the AAAI Conference on Artificial Intelligence},
  volume={39},
  number={25},
  pages={26642--26650},
  year={2025}
}

@inproceedings{zhou2025code,
  title={Code-as-monitor: Constraint-aware visual programming for reactive and proactive robotic failure detection},
  author={Zhou, Enshen and Su, Qi and Chi, Cheng and Zhang, Zhizheng and Wang, Zhongyuan and Huang, Tiejun and Sheng, Lu and Wang, He},
  booktitle={Proceedings of the Computer Vision and Pattern Recognition Conference},
  pages={6919--6929},
  year={2025}
}

@article{xu2025affordance,
  title={Affordance Field Intervention: Enabling VLAs to Escape Memory Traps in Robotic Manipulation},
  author={Xu, Siyu and Wang, Zijian and Wang, Yunke and Xia, Chenghao and Huang, Tao and Xu, Chang},
  journal={arXiv preprint arXiv:2512.07472},
  year={2025}
}

@article{yang2025fpc,
  title={FPC-VLA: A Vision-Language-Action Framework with a Supervisor for Failure Prediction and Correction},
  author={Yang, Yifan and Duan, Zhixiang and Xie, Tianshi and Cao, Fuyu and Shen, Pinxi and Song, Peili and Jin, Piaopiao and Sun, Guokang and Xu, Shaoqing and You, Yangwei and others},
  journal={arXiv preprint arXiv:2509.04018},
  year={2025}
}

@article{lin2025failsafe,
  title={Failsafe: Reasoning and recovery from failures in vision-language-action models},
  author={Lin, Zijun and Duan, Jiafei and Fang, Haoquan and Fox, Dieter and Krishna, Ranjay and Tan, Cheston and Wen, Bihan},
  journal={arXiv preprint arXiv:2510.01642},
  year={2025}
}

@article{carion2025sam,
  title={Sam 3: Segment anything with concepts},
  author={Carion, Nicolas and Gustafson, Laura and Hu, Yuan-Ting and Debnath, Shoubhik and Hu, Ronghang and Suris, Didac and Ryali, Chaitanya and Alwala, Kalyan Vasudev and Khedr, Haitham and Huang, Andrew and others},
  journal={arXiv preprint arXiv:2511.16719},
  year={2025}
}

@article{intelligence2025pi,
  title={{$\pi_{0.6}^{*}$: a VLA That Learns From Experience}},
  author={Intelligence, Physical and Amin, Ali and Aniceto, Raichelle and Balakrishna, Ashwin and Black, Kevin and Conley, Ken and Connors, Grace and Darpinian, James and Dhabalia, Karan and DiCarlo, Jared and others},
  journal={arXiv preprint arXiv:2511.14759},
  year={2025}
}

@article{akaike1974new,
  title={A new look at the statistical model identification},
  author={Akaike, Hirotugu},
  journal={IEEE transactions on automatic control},
  volume={19},
  number={6},
  pages={716--723},
  year={1974},
  publisher={Ieee}
}

@article{massey1951kolmogorov,
  title={The Kolmogorov-Smirnov test for goodness of fit},
  author={Massey Jr, Frank J},
  journal={Journal of the American statistical Association},
  volume={46},
  number={253},
  pages={68--78},
  year={1951},
  publisher={Taylor \& Francis}
}

@article{lin2025depth,
  title={Depth anything 3: Recovering the visual space from any views},
  author={Lin, Haotong and Chen, Sili and Liew, Junhao and Chen, Donny Y and Li, Zhenyu and Shi, Guang and Feng, Jiashi and Kang, Bingyi},
  journal={arXiv preprint arXiv:2511.10647},
  year={2025}
}

@inproceedings{luo2024serl,
  title={Serl: A software suite for sample-efficient robotic reinforcement learning},
  author={Luo, Jianlan and Hu, Zheyuan and Xu, Charles and Tan, You Liang and Berg, Jacob and Sharma, Archit and Schaal, Stefan and Finn, Chelsea and Gupta, Abhishek and Levine, Sergey},
  booktitle={2024 IEEE International Conference on Robotics and Automation (ICRA)},
  pages={16961--16969},
  year={2024},
  organization={IEEE}
}

@inproceedings{vats2023efficient,
  title={Efficient recovery learning using model predictive meta-reasoning},
  author={Vats, Shivam and Likhachev, Maxim and Kroemer, Oliver},
  booktitle={2023 IEEE International Conference on Robotics and Automation (ICRA)},
  pages={7258--7264},
  year={2023},
  organization={IEEE}
}

@article{zhang2026ma,
  title={MA-VLA: Multi-Arm Vision-Language-Action Model for Collaboration and Compositional Generalization},
  author={Zhang, Zaibin and Xiao, Junlan and Zhang, Zhongbo and Wang, Yifan and Kang, Li and Qin, Yiran and Xia, Changxing and Zhou, Heng and Fu, Talas and Zhou, Enshen and others},
  journal={arXiv preprint arXiv:2608.25864},
  year={2026}
}

\clearpage

\twocolumn[
\begin{center}
    \vspace{0.25in}
    {\LARGE Supplementary Material\par}
    \vspace{1.0em}
\end{center}
]

\section{Overview}
\label{sec:sup}

This supplement provides implementation details, benchmark specifications,
additional analyses, training protocols, and qualitative results that complement
the main paper. It is organized as follows:

\begin{itemize}
    \item \textbf{Sec.~\ref{sec:data}: Experience-Guided Corrective Data Synthesis Details.}
    Additional details on stage-conditioned failure modeling, distribution fitting,
    and corrective-state synthesis.

    \item \textbf{Sec.~\ref{sec:atomic}: Atomic Corrective Execution Details.}
    Implementation details of the offline VLM planner, event-triggered 3D monitor,
    and geometric predicates.

    \item \textbf{Sec.~\ref{sec:benchmark}: FSR-Bench Details.}
    Recovery horizons, failure-state taxonomy, train/test distribution construction,
    and evaluation protocol.

    \item \textbf{Sec.~\ref{sec:additional_experiment}: Additional Experimental Analysis.}
    Additional analysis of corrective interventions, the number of experience-guided
    rollouts, rollback recovery, and monitor accuracy.

    \item \textbf{Sec.~\ref{sec:training}: Training and Evaluation Protocols.}
    Training hyperparameters, data budgets, deployment details, and evaluation settings.

    \item \textbf{Sec.~\ref{sec:visualization}: Visualization.}
    Qualitative visualizations of nominal and corrective execution for
    \textit{Place Dual Shoes}, together with all 36 FSR-Bench failure states.
\end{itemize}

\section{Experience-Guided Corrective Data Synthesis Details}
\label{sec:data}

As described in the main paper, CARE begins by executing the nominal atomic plans
without corrective supervision and collecting failures from 100 preliminary rollouts.
A failure is recorded when an atomic stage does not satisfy its termination condition.
For a failure at stage $k_i$, we record the stage-critical relative deviation
\begin{equation}
    d_i = (\Delta \tau_i,\Delta R_i),
\end{equation}
where $\Delta \tau_i\in\mathbb{R}^3$ is the translational deviation and
$\Delta R_i$ is the rotational deviation. For grasping, the deviation is measured
between the gripper and the manipulated object at gripper closure. For placement,
it is measured between the object and the target at gripper opening. We therefore
model a stage-conditioned failure prior $p(d\mid k)$ rather than a single
task-level perturbation distribution.

For implementation, we parameterize the translational component by
$(dx,dy,dz)$. In the tasks used in our experiments, the dominant rotational
variation around grasping and placement is yaw, so the rotational component is
represented by a scalar $dr$ in the distribution-fitting analysis. This is the
four-variable representation used in the rollout-stability experiments in
Sec.~\ref{sec:rollout_num}.

For each scalar deviation variable, we consider a small family of parametric
distributions, including Gaussian, Beta, Gamma, Weibull, Log-Normal, and Uniform
distributions. We select the best-fitting family using goodness-of-fit statistics
together with the Akaike Information Criterion (AIC). This procedure keeps the
corrective-data prior grounded in failures encountered by the policy instead of
assuming a manually chosen perturbation shape.

To synthesize a corrective trajectory at stage $k$, we sample
\begin{equation}
    d=(\Delta\tau,\Delta R)\sim p(d\mid k),
\end{equation}
perturb the stage-relevant relative pose,
\begin{equation}
    \tau_k \leftarrow \tau_k+\Delta\tau,\qquad
    R_k \leftarrow \Delta R R_k,
\end{equation}
and then roll the system forward under the simulator or real-world physical
dynamics. The resulting state, rather than the injected pose itself, is used as
the start state for expert correction. Consequently, synthesized failures are
new physical states generated from the learned failure prior rather than direct
replays of previously collected failures.

Corrective demonstrations are collected only for the recovery behavior needed
at the current atomic stage. For example, a re-grasp trajectory terminates once
a stable grasp has been recovered instead of continuing to the end of the full
task. This short-horizon design reduces corrective-data collection cost while
preserving the supervision required to learn the corrective atomic behavior.

\paragraph{Relation to FSR-Bench.}
The experience-guided distribution above is used for corrective data synthesis
in the standard RoboTwin, RoboFactory, and real-world CARE experiments. FSR-Bench
uses a deliberately different protocol to avoid coupling recovery training to the
test distribution: its recovery training trajectories are collected from
\emph{uniformly sampled} perturbations within the predefined range of each recovery
scenario, whereas evaluation initializations are sampled from empirical
stage-conditioned failure distributions estimated from nominal-policy rollouts.
We detail this separation in Sec.~\ref{sec:benchmark}.

\section{Atomic Corrective Execution Details}
\label{sec:atomic}

\subsection{Offline VLM planner}

The planner is implemented with GPT-4.1 at temperature $0.0$ and is queried once
at task initialization. Given the high-level instruction and the initial global-view
image, it decomposes the task into $K$ atomic stages. Consistent with the notation
of the main paper, each stage plan is
\begin{equation}
    \Pi_k=(u_k,\gamma_k,\psi_k,\mathcal{U}_k),
    \qquad
    \mathcal{U}_k=\mathcal{U}^{\mathrm{adj}}_k\cup
    \mathcal{U}^{\mathrm{re}}_k,
\end{equation}
where $u_k$ is the nominal atomic instruction, $\gamma_k$ specifies the geometric
cues used by the monitor, $\psi_k$ is the stage-termination condition, and
$\mathcal{U}^{\mathrm{adj}}_k$ and $\mathcal{U}^{\mathrm{re}}_k$ contain
intra-stage adjustment and stage-level re-operation instructions, respectively.
The resulting stage plans remain fixed during low-level execution.

\begin{promptbox}{Prompt Template for the Offline VLM Planner}
You are an offline VLM planner for CARE (Corrective Atomic Robotic Execution)
in multi-arm manipulation.

Your task is to decompose a high-level manipulation instruction into ordered
atomic stages and produce the stage plan used by the 3D monitor and VLA executor.

Inputs:
- {global image}
- {task instruction}
- {task description}
- {agent description}
- {atomic action pool}
- {deviation statistics}

Requirements:
1. Decompose the task into ordered atomic stages.
2. For each stage, assign exactly one nominal atomic instruction to each agent.
3. For each stage, provide geometric guidance (*@$\gamma_k$@*) describing the
   task-relevant gripper/object/target relations that should be monitored.
4. For each stage, provide a concise observable termination condition (*@$\psi_k$@*).
5. Allocate corrective instructions only for deviation cases relevant to the stage.
6. Separate corrective instructions into:
   - U_k_adj: intra-stage adjustments for local deviations;
   - U_k_re: stage-level re-operations such as re-grasp or re-place.
7. Use only actions from the atomic action pool.
8. Use "Wait for the other robots." for idle agents.
9. Ensure physical feasibility, spatial consistency, and coordinated multi-arm execution.
10. Output only the Python dictionary.

Planning principles:
- Decompose the task according to object dependencies and inter-arm coordination.
- Respect geometry and reachability when allocating actions to agents.
- Grasp must happen before lift; lift must happen before transfer or placement.
- Synchronize agents within the same stage when cooperative manipulation is required.
- Use U_k_adj for deviations that can be corrected before the stage-critical event.
- Use U_k_re when the stage outcome has failed and the atomic operation must be repeated.
- The termination condition should depend on observable progress in the current stage.

Output format:
stage_plans = {
    "stage 1": {
        "u_k": {
            "agent0": "<nominal atomic action>",
            "agent1": "<nominal atomic action>",
            ...
        },
        "gamma_k": "<geometric guidance for monitoring>",
        "psi_k": "<stage termination condition>",
        "U_k_adj": {
            "<deviation case 1>": {
                "agent0": "<corrective atomic action>",
                "agent1": "<corrective atomic action>",
                ...
            }
        },
        "U_k_re": {
            "<failure case 1>": {
                "agent0": "<re-operation atomic action>",
                "agent1": "<re-operation atomic action>",
                ...
            }
        }
    },
    ...
}
\end{promptbox}
\subsection{Event-triggered 3D geometric monitor}

\begin{figure*}[t]
    \centering
    \includegraphics[width=\textwidth]{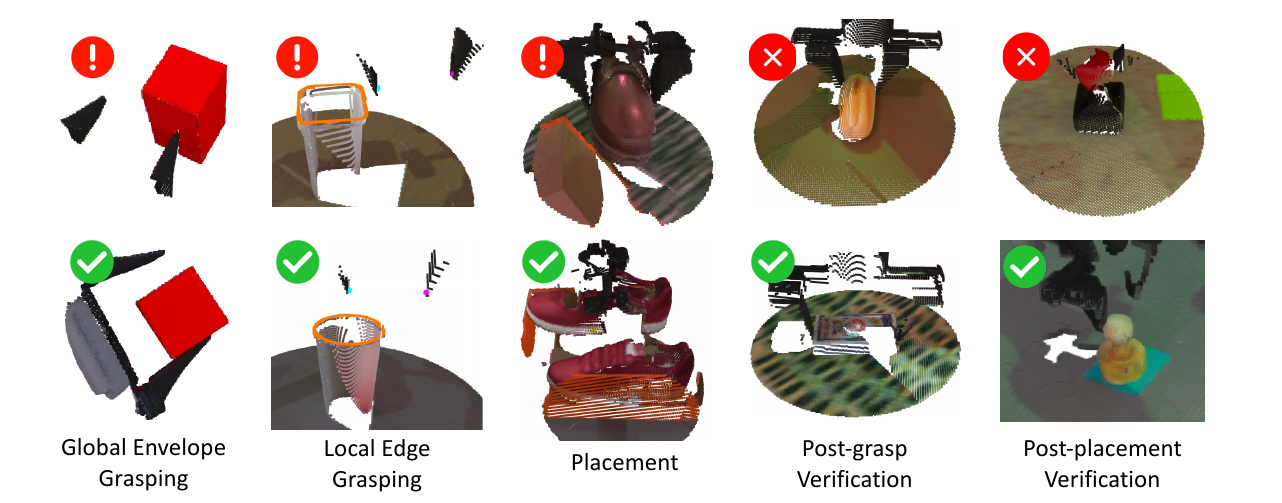}
    \caption{Visualization of point-cloud-based geometric judgement.
    \protect\raisebox{-0.1em}{\includegraphics[height=1em]{figs/wrong.png}}
    indicates cases requiring post-execution re-operation;
    \protect\raisebox{-0.1em}{\includegraphics[height=1em]{figs/adj.png}}
    indicates cases requiring intra-execution adjustment; and
    \protect\raisebox{-0.1em}{\includegraphics[height=1em]{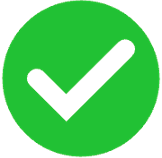}}
    indicates correct cases.}
    \label{fig:pointcloud}
\end{figure*}

The 3D monitor is a point-cloud predicate checker rather than a learned recovery
policy or a VLM-based judge. During execution, CARE continuously tracks the
gripper open/close state, but geometric perception is invoked only at
stage-critical transitions (e.g., grasp closure or object release) or shortly
before a predicted transition when intra-execution adjustment is enabled.
This event-triggered design matches the execution protocol used in the main paper.

At each triggered check, SAM 3~\cite{carion2025sam} extracts masks for the gripper,
manipulated object, and target region, while Depth Anything 3
(DA3)~\cite{lin2025depth} provides depth. We fuse these observations into
point clouds $\mathcal{P}_{g,t}$, $\mathcal{P}_{o,t}$, and
$\mathcal{P}_{\mathrm{tar},t}$ and compute a compact geometric state
\begin{equation}
    q_t=\Phi(\mathcal{P}_{g,t},\mathcal{P}_{o,t},\mathcal{P}_{\mathrm{tar},t}).
\end{equation}
Conditioned on the active stage plan $\Pi_{k_t}$, the monitor selects
\begin{equation}
    \tilde{u}_t=f_{\mathrm{mon}}(\Pi_{k_t},q_t),\qquad
    \tilde{u}_t\in
    \{u_{k_t}\}\cup
    \mathcal{U}^{\mathrm{adj}}_{k_t}\cup
    \mathcal{U}^{\mathrm{re}}_{k_t}.
\end{equation}
The same VLA executor subsequently executes nominal, adjustment, and re-operation
instructions.

The monitor uses skill-level geometric predicate templates rather than
task-specific learned classifiers. Once a predicate template is defined for an
atomic skill, it is reused across tasks; only the stage-relevant geometric cues
and numerical tolerances are supplied through $\gamma_k$ and calibrated from
rollout statistics.

\paragraph{Intra-execution adjustment.}
During grasping and approach, the monitor estimates the approach vector
$\mathbf{v}_{\mathrm{app}}$ from the recent TCP trajectory and projects the local
geometry onto the orthogonal plane $\mathcal{H}_{\perp}$. For global envelope
grasping, let $\tilde{p}_{\mathrm{tcp}}$ and
$\tilde{\mathcal{L}}_{\mathrm{axis}}$ denote the projected TCP position and object
principal axis. Adjustment is triggered when
\begin{equation}
    \phi_{\mathrm{geg}}
    =
    \mathbb{I}
    \left[
        d(\tilde{p}_{\mathrm{tcp}},
        \tilde{\mathcal{L}}_{\mathrm{axis}})
        >
        \delta_{\mathrm{geg}}
    \right].
\end{equation}
For local edge grasping, let $\tilde{S}$ be the projected line segment connecting
the two fingertips and $\partial\tilde{\mathcal{P}}_o$ the projected object
boundary. We trigger adjustment if the fingertips fail to straddle the edge or
violate the safety clearance:
\begin{equation}
    \phi_{\mathrm{leg}}
    =
    \mathbb{I}
    \left[
        (\tilde{S}\cap\partial\tilde{\mathcal{P}}_o=\emptyset)
        \vee
        \left(
        \min_{i\in\{1,2\}}
        d(\tilde{p}_i,\partial\tilde{\mathcal{P}}_o)
        \leq
        \delta_{\mathrm{safe}}
        \right)
    \right].
\end{equation}
For placement, both the effective grasp center $p_{\mathrm{gc}}$ and target
position $p_{\mathrm{target}}$ are projected onto the support plane:
\begin{equation}
    \phi_{\mathrm{place}}
    =
    \mathbb{I}
    \left[
        \left\|
        \Pi_{\mathcal{H}_{\mathrm{support}}}(p_{\mathrm{gc}})
        -
        \Pi_{\mathcal{H}_{\mathrm{support}}}(p_{\mathrm{target}})
        \right\|_2
        >
        \delta_{\mathrm{place}}
    \right].
\end{equation}
If any corresponding adjustment predicate is active, CARE selects an instruction
from $\mathcal{U}^{\mathrm{adj}}_{k_t}$.

\paragraph{Post-stage re-operation.}
After a stage-critical event, CARE performs discrete verification of the stage
outcome. For grasp verification, let $p_{\mathrm{obj}}$ denote the object center,
$p_{\mathrm{gc}}$ the grasp center, $\mathbf{n}_1$ and $\mathbf{n}_2$ the surface
normals at the two contact points, and $w_{\mathrm{gripper}}$ the measured
gripper width. Re-grasp is triggered when
\begin{equation}
\begin{aligned}
\phi_{\mathrm{re\text{-}grasp}}
= \mathbb{I}\Big[
&\|p_{\mathrm{gc}}-p_{\mathrm{obj}}\|_2
> \delta_{\mathrm{center}}
\ \lor \\
&\mathbf{n}_1^\top \mathbf{n}_2 \geq 0
\ \lor \\
&w_{\mathrm{gripper}}
\leq \epsilon_{\mathrm{empty}}
\Big].
\end{aligned}
\label{eq:regrasp}
\end{equation}

For post-placement verification, re-placement is triggered when
\begin{equation}
\begin{aligned}
\phi_{\mathrm{re\text{-}place}}
= \mathbb{I}\Big[
&\big\|
\Pi_{\mathcal{H}_{\mathrm{support}}}
(p_{\mathrm{obj}}^{\mathrm{final}})
-
\Pi_{\mathcal{H}_{\mathrm{support}}}
(p_{\mathrm{target}})
\big\|_2 \\
&> \delta_{\mathrm{re\text{-}place}}
\Big].
\end{aligned}
\label{eq:replace}
\end{equation}

When either re-operation predicate is active, CARE selects an instruction from
$\mathcal{U}^{\mathrm{re}}_{k_t}$.

The geometric tolerances
$\delta_{\mathrm{geg}}$, $\delta_{\mathrm{safe}}$,
$\delta_{\mathrm{place}}$, $\delta_{\mathrm{center}}$, and
$\delta_{\mathrm{re\text{-}place}}$
are calibrated from stage-conditioned rollout statistics rather than tuned as
task-specific learned classifiers. The empty-grasp threshold
$\epsilon_{\mathrm{empty}}$ is determined by the mechanical closing limit of
the parallel-jaw gripper.

\section{FSR-Bench Details}
\label{sec:benchmark}

FSR-Bench evaluates recovery independently from nominal task execution. Each
episode begins from an intermediate failure state and asks the policy to reach
the scenario-specific recovery target within a predefined horizon. The benchmark
contains 36 recovery scenarios across five tasks: 21 \emph{Easy} local failures
that typically admit one corrective operation and 15 \emph{Hard} structural
failures that require multi-step correction or coordinated multi-arm recovery.

\begin{table}[t]
\centering
\caption{Maximum recovery steps for each task in FSR-Bench.}
\label{tab:max_steps}
\footnotesize
\setlength{\tabcolsep}{5pt}
\renewcommand{\arraystretch}{1.05}
\begin{tabularx}{\columnwidth}{@{}>{\raggedright\arraybackslash}Xcc@{}}
\toprule
\textbf{Task} & \textbf{Easy} & \textbf{Hard} \\
\midrule
Pot Lift Recovery            & 100 & 400 \\
Skillet Placement Recovery   & 200 & 250 \\
Block Handover Recovery      & 400 & 500 \\
Seal Stamp Recovery          & 200 & 500 \\
Dual Shoes Recovery          & 200 & 400 \\
\bottomrule
\end{tabularx}
\end{table}

\begin{table}[t]
\centering
\caption{Failure state types and corresponding tasks in FSR-Bench. DSR: Dual Shoes Recovery; SSR: Seal Stamp Recovery; SPR: Skillet Placement Recovery; BHR: Block Handover Recovery; PLR: Pot Lift Recovery.}
\label{tab:failure_types}
\footnotesize
\renewcommand{\arraystretch}{1.0}
\begin{tabular*}{\columnwidth}{@{\extracolsep{\fill}} l p{0.65\columnwidth}}
\toprule
Failure State Type & Tasks \\
\midrule
Pose Misalignment & BHR\_easy, DSR\_easy \\
Grasp Failure & SSR\_easy, BHR\_easy, PLR\_easy, SPR\_easy, DSR\_easy \\
Placement Offset & SSR\_easy, DSR\_easy \\
Tipping & SSR\_hard, PLR\_hard, SPR\_hard, DSR\_hard \\
Unreachable Target & SSR\_hard, BHR\_hard \\
Target Occupation & SSR\_hard, BHR\_hard \\
Accidental Drop &  SPR\_hard \\
Carry-over Error & PLR\_hard, DSR\_hard \\
\bottomrule
\end{tabular*}
\end{table}

Table~\ref{tab:max_steps} reports the maximum recovery horizon used for each task
and regime. The larger hard-regime budgets reflect the additional atomic
operations often required by structural anomalies.

Table~\ref{tab:failure_types} summarizes the failure taxonomy. \textit{Pose
Misalignment} denotes gripper/object or object/target spatial misalignment;
\textit{Grasp Failure} includes missed, unstable, or slipping grasps;
\textit{Placement Offset} denotes an incorrect final placement;
\textit{Tipping} denotes loss of object stability; \textit{Unreachable Target}
places the desired target outside the immediately feasible workspace;
\textit{Target Occupation} blocks the intended target region with another object;
\textit{Accidental Drop} is unintended release during manipulation;
\textit{Mis-stacked Objects} violates the intended stacking relation; and
\textit{Carry-over Error} denotes a failure from an earlier stage that remains
unresolved and affects subsequent execution. In \textit{Dual Shoes
Recovery\_hard}, for example, an incorrectly oriented first shoe can remain in
the scene while the second placement also fails, yielding a multi-stage recovery
problem.

\subsection{Training and test distributions}

FSR-Bench deliberately separates the recovery-training distribution from the
test distribution. For recovery training, perturbations are sampled
\emph{uniformly} within the predefined range of each recovery scenario, and all
backbone variants use the same 50 corrective trajectories per recovery type.

For evaluation, we first execute nominal policies without recovery and estimate
a stage-conditioned empirical failure distribution
\begin{equation}
    p_{\mathrm{fail}}(d\mid k).
\end{equation}
At test time, we sample $d\sim p_{\mathrm{fail}}(d\mid k)$, inject the deviation
at the corresponding stage of a nominal execution, and roll the system forward
under physical dynamics. The state after this rollout becomes the actual recovery
initialization. Therefore, test episodes are newly generated failure states
grounded in observed policy failures rather than memorized replays or direct
copies of training perturbations.

Evaluation episodes additionally randomize object geometry, table texture,
lighting, distractor objects, and object placement. For a given recovery scenario,
all methods are evaluated on the same pre-generated failure initializations.

\subsection{Evaluation protocol}

\begin{figure*}[t]
    \centering
    \includegraphics[width=\textwidth]{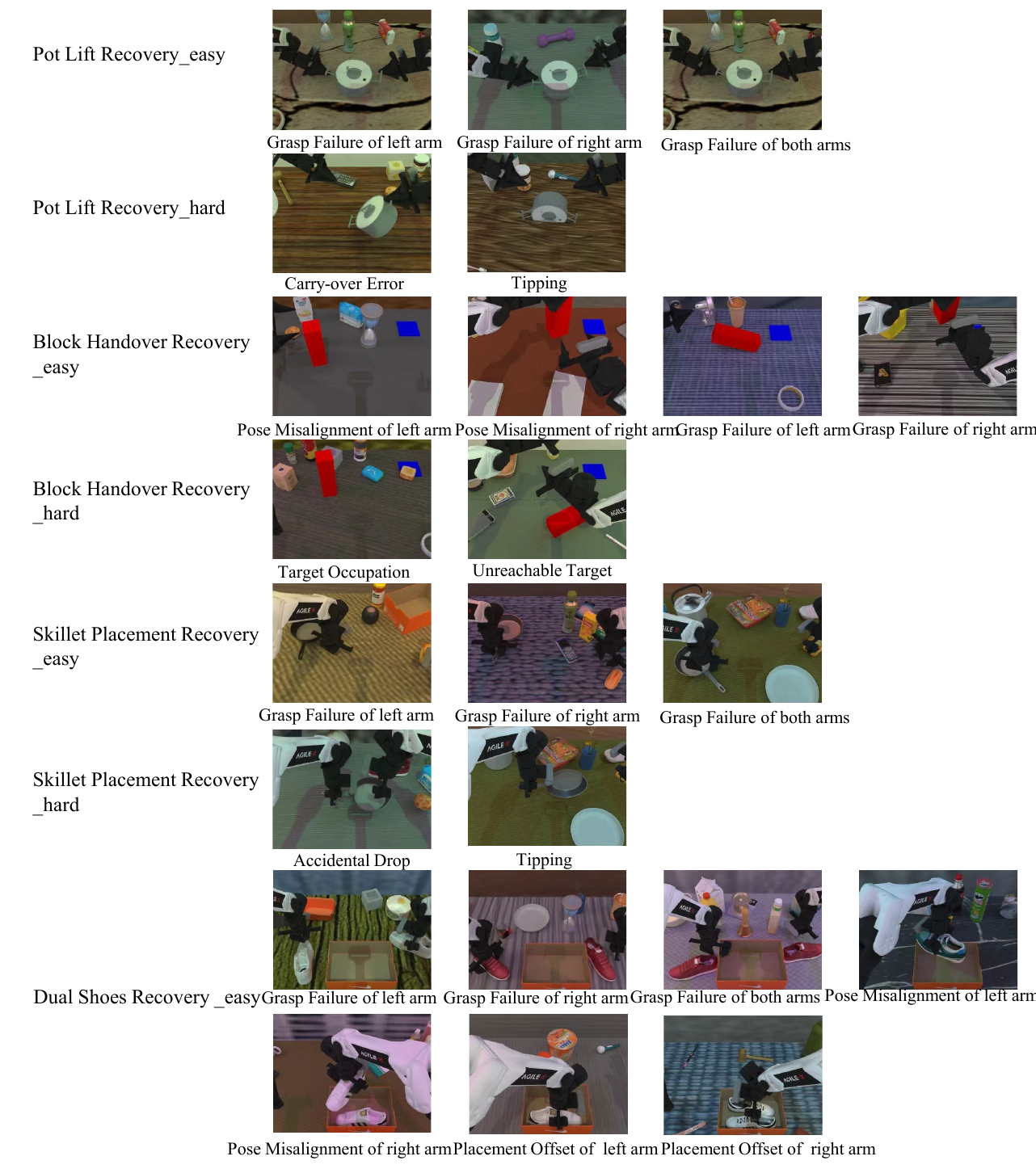}
    \caption{Visualization of the 36 failure states in FSR-Bench (Part 1/2). The benchmark covers 21 easy and 15 hard cases spanning local deviations and structural failures.}
    \label{fig:fsr_bench_1}
\end{figure*}

\begin{figure*}[t]
    \centering
    \includegraphics[width=\textwidth]{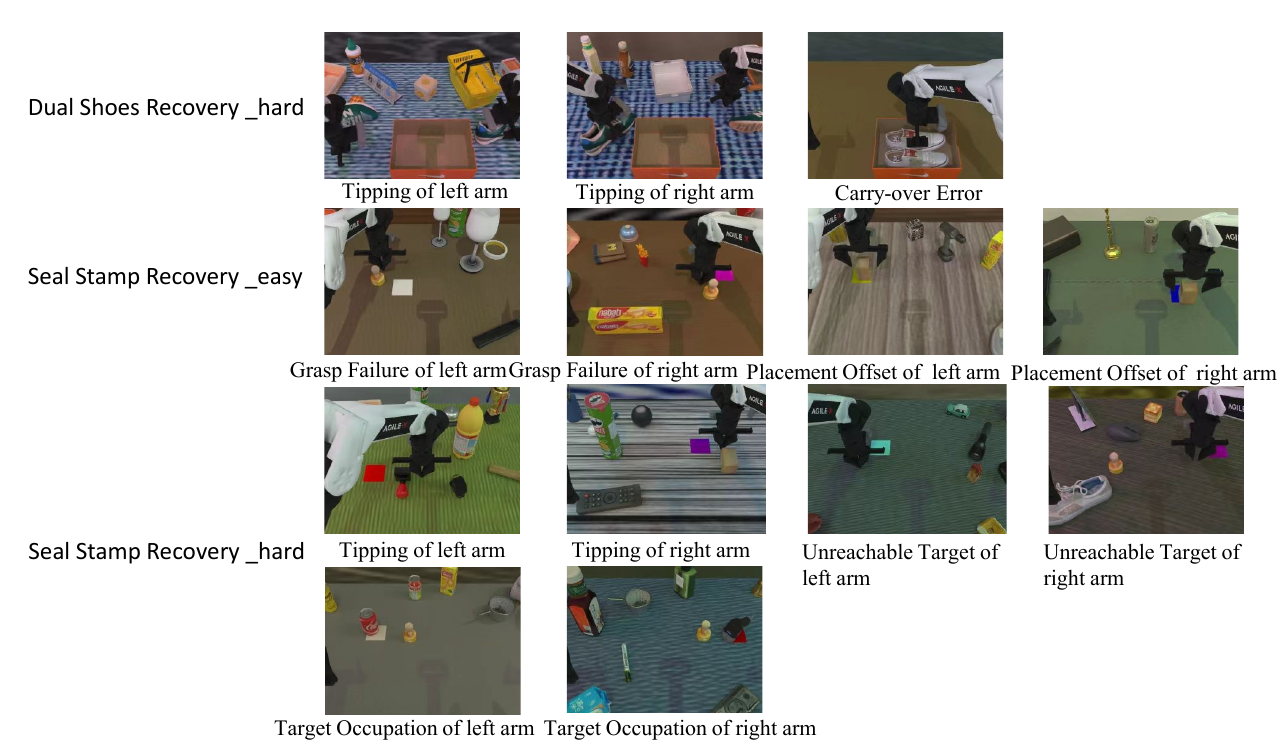}
    \caption{Visualization of the 36 failure states in FSR-Bench (Part 2/2).}
    \label{fig:fsr_bench_2}
\end{figure*}

Each method is evaluated over $N_{\mathrm{eval}}=100$ trials per
task--regime pair, with initializations sampled from the corresponding
constituent recovery scenarios. Recovery Success Rate (RSR) is
\begin{equation}
    \mathrm{RSR}
    =
    \frac{1}{N_{\mathrm{eval}}}
    \sum_{j=1}^{N_{\mathrm{eval}}}\mathbb{I}[r_j=1],
\end{equation}

where $r_j=1$ if the policy reaches the scenario-specific recovery
target within its prescribed horizon, and $r_j=0$ otherwise.
For each scenario, the recovery target and horizon are shared across
methods. For each task and regime, RSR is computed over 100 trials
sampled from its constituent failure states, using the same
pre-generated initializations across methods.Figures~\ref{fig:fsr_bench_1} and~\ref{fig:fsr_bench_2} visualize all
36 failure states.

\section{Additional Experimental Analysis}
\label{sec:additional_experiment}

\subsection{Analysis of corrective interventions}

\begin{figure*}[t]
    \centering
    \includegraphics[width=\textwidth]{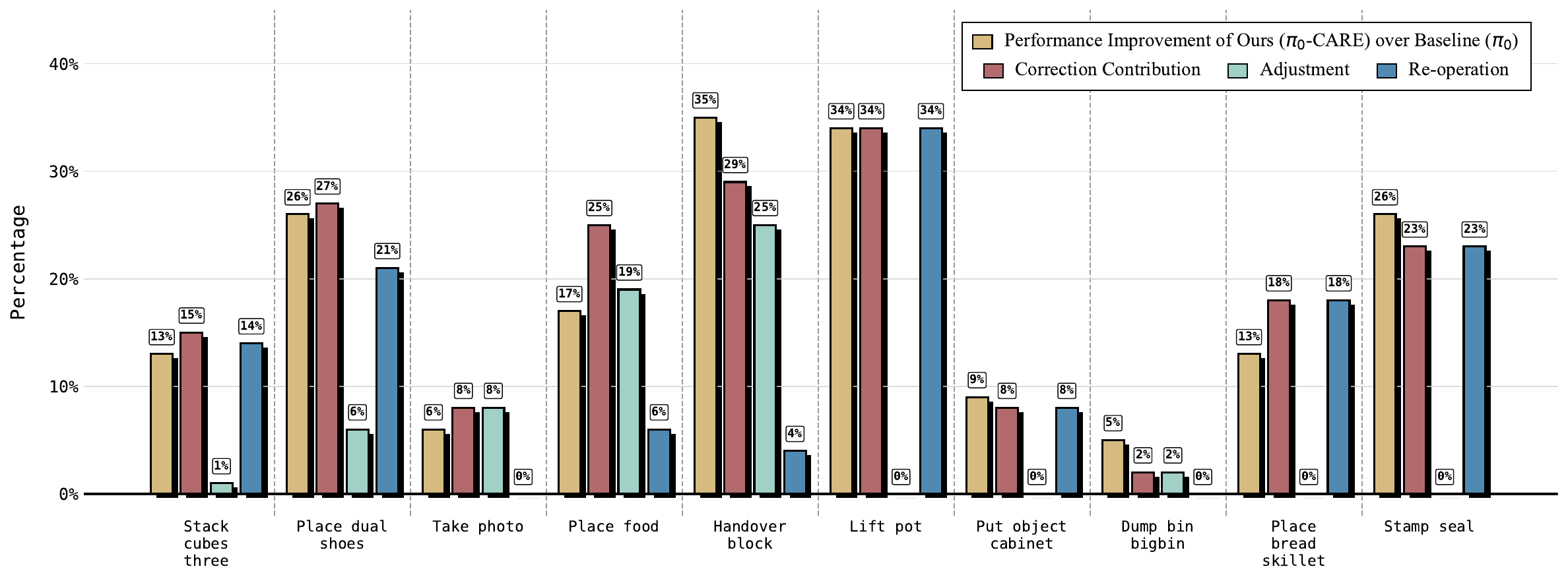}
    \caption{\textbf{Breakdown of successful corrective interventions for
    $\bm{\pi}_0$.} \textit{Adjustment} denotes successful intra-execution
    corrections, while \textit{Re-operation} denotes successful repeated atomic
    operations after a failed stage outcome.}
    \label{fig:gain}
\end{figure*}

Figure~\ref{fig:gain} provides an intervention-level view complementary to the
component ablation in the main paper. The component ablation separates the effect
of corrective training data from the effect of atomic corrective execution,
whereas Fig.~\ref{fig:gain} focuses only on what happens during evaluation after
the full CARE model has been trained. Successful recovery episodes include both
intra-execution adjustments and post-stage re-operations, confirming that the two
corrective primitives are both used by the closed-loop system. We therefore do
not interpret this figure as replacing the Data/Execution component ablation;
rather, it illustrates how the learned corrective behaviors are invoked online.

\subsection{Event-triggered monitoring protocol}

The VLM-monitor baseline and our 3D monitor use the same triggered observations
in the monitoring ablation of the main paper. A check is invoked at stage-critical
events such as gripper closure or object release. When a stage supports
intra-execution adjustment, CARE additionally checks shortly before a predicted
critical transition; the prediction window follows the VLA action horizon of 50.
For stages whose boundaries are not directly associated with a gripper-state
change, the system performs a check once per 50 executed actions. Using the same
trigger schedule isolates the quality and latency of the monitoring mechanism
itself rather than differences in invocation frequency.

\subsection{Ablation on the Number of Experience-Guided Rollouts}
\label{sec:rollout_num}

We ablate the number of preliminary rollouts used to estimate the
experience-guided failure distribution. We compare 50, 100, and 150 rollouts and
keep the amount of corrective supervision fixed at 50 synthesized corrective
atomic trajectories per error type. Table~\ref{tab:ablation_rollout_num} reports
three simulation tasks whose 100-rollout setting is identical to the main-paper
training protocol.

Across these tasks, 50 rollouts are consistently weaker, whereas the difference
between 100 and 150 rollouts is small. The average success rate is 45.0\% with
50 rollouts, 51.5\% with 100 rollouts, and 51.0\% with 150 rollouts. This supports
the use of 100 preliminary rollouts as a practical trade-off between collection
cost and the stability of the empirical failure prior.

To analyze the prior directly, we examine \textit{Put Object Cabinet} using the
four modeled variables $dx$, $dy$, $dz$, and $dr$. We compare the distributions
estimated from prefixes of 50, 100, 150, and 200 rollouts and track cumulative
mean/variance, empirical PDFs, and Jensen--Shannon (JS) divergence to the
full-data distribution. At 50 rollouts, the JS divergences are 0.0982, 0.1422,
0.0582, and 0.1556 for $dx$, $dy$, $dz$, and $dr$, respectively. At 100 rollouts,
they drop to 0.0224, 0.0200, 0.0374, and 0.0245; the cumulative statistics also
become substantially flatter. The 100-rollout setting provides 92 valid samples
for each translational variable and 63 samples for $dr$. Increasing the budget to
150 rollouts only further reduces the JS divergences to 0.0145, 0.0113, 0.0141,
and 0.0174. These results indicate that the failure prior has largely stabilized
by 100 rollouts.

\begin{figure}[t]
    \centering
    \includegraphics[width=\linewidth]{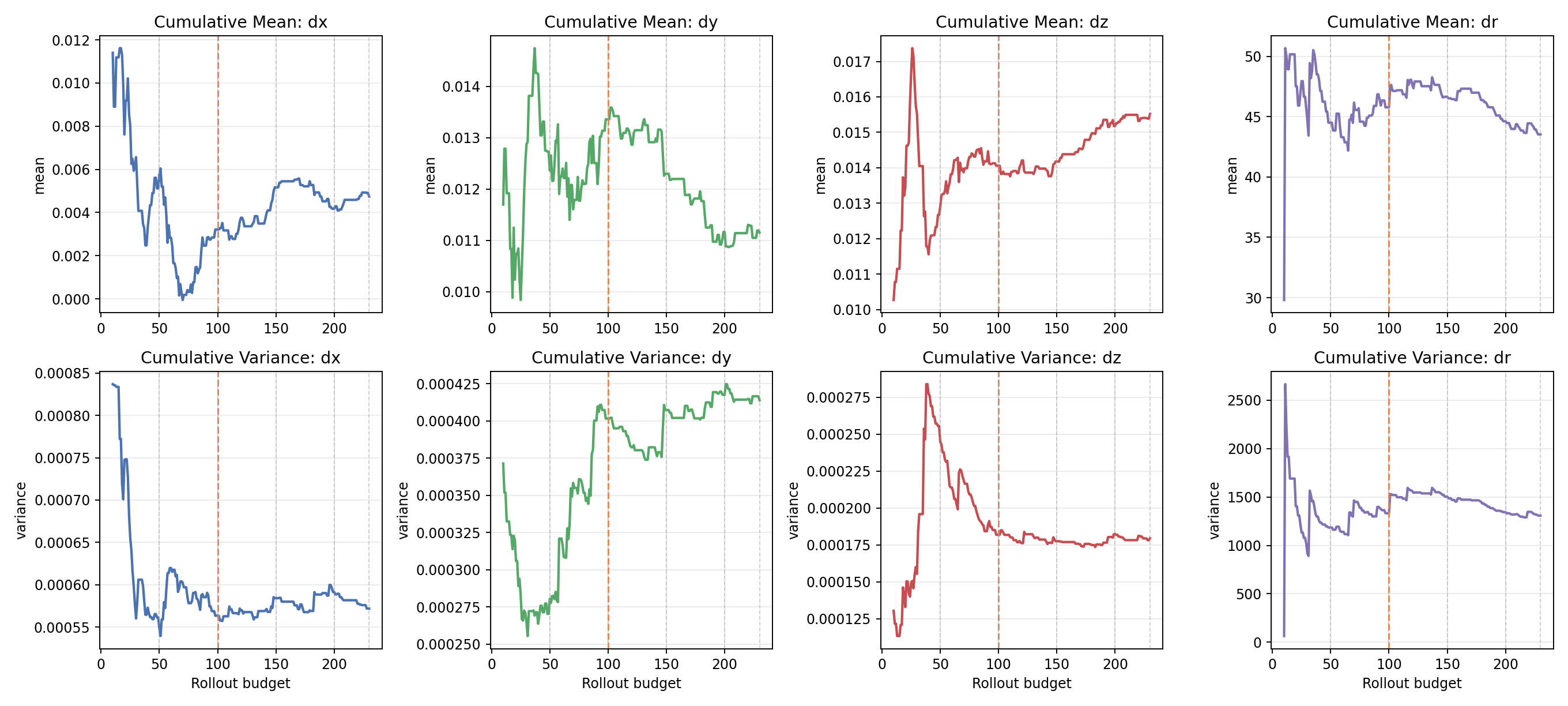}
    \caption{Cumulative mean and variance of the four modeled variables, $dx$, $dy$, $dz$, and $dr$, under different rollout budgets on \textit{Put Object Cabinet}. The statistics become much more stable around 100 rollouts, while further increasing the rollout budget yields only limited changes.}
    \label{fig:rollout_running_stats_put_object_cabinet}
\end{figure}

\begin{figure}[t]
    \centering
    \includegraphics[width=\linewidth]{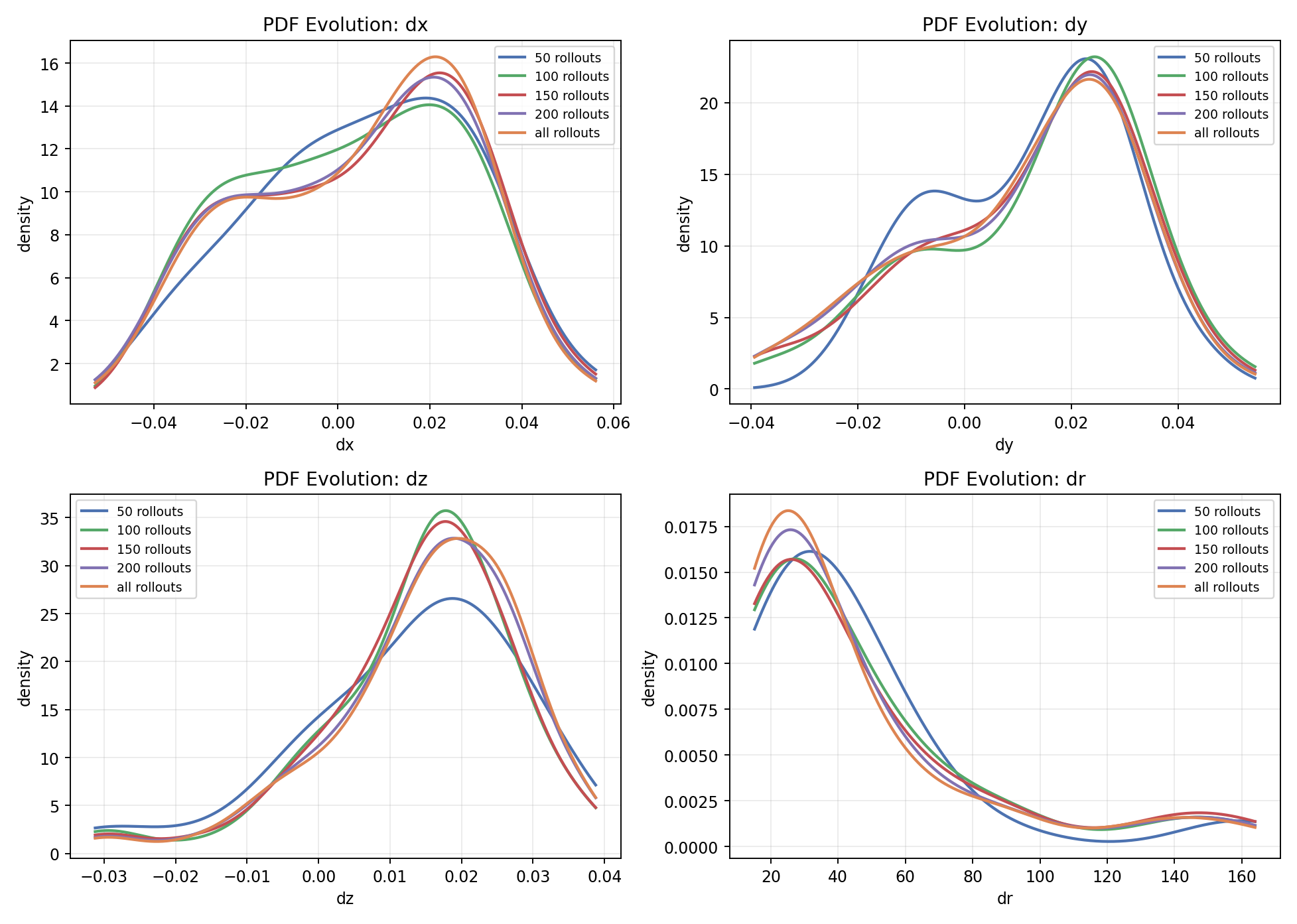}
    \caption{Evolution of the empirical PDFs of $dx$, $dy$, $dz$, and $dr$ under different rollout budgets on \textit{Put Object Cabinet}. The distributions estimated from 100 rollouts already closely match those obtained with larger rollout budgets.}
    \label{fig:rollout_pdf_put_object_cabinet}
\end{figure}

\begin{figure}[t]
    \centering
    \includegraphics[width=\linewidth]{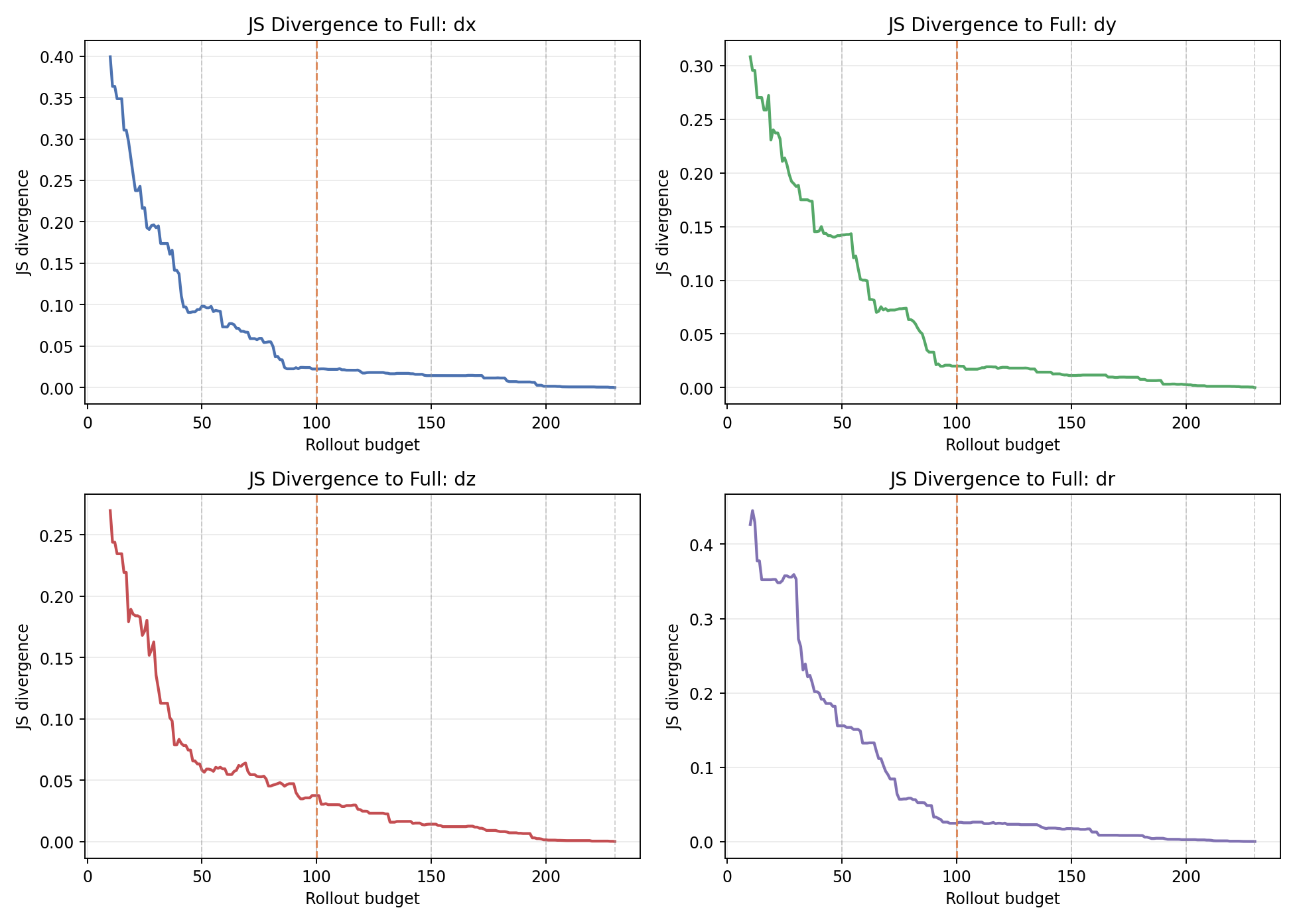}
    \caption{Jensen-Shannon divergence between the prefix distribution and the full-data distribution for $dx$, $dy$, $dz$, and $dr$ under different rollout budgets on \textit{Put Object Cabinet}. The divergence drops substantially by 100 rollouts and decreases only marginally afterward, indicating that the deviation prior has largely saturated.}
    \label{fig:rollout_js_put_object_cabinet}
\end{figure}

\begin{table}[t]
\centering
\caption{Ablation on the number of experience rollouts used for corrective data synthesis.
The amount of corrective supervision is fixed across all settings.}
\label{tab:ablation_rollout_num}
\footnotesize
\setlength{\tabcolsep}{4pt}
\renewcommand{\arraystretch}{1.1}
\resizebox{\columnwidth}{!}{%
\begin{tabular}{llccc}
\toprule
Task & Backbone & 50 & 100 & 150 \\
\midrule
\multirow{2}{*}{Lift Pot}
& $\pi_0$      & 83\% & \textbf{91\%} & 87\% \\
& $\pi_0$-FAST & 31\% & \textbf{40\%} & 36\% \\
\midrule
\multirow{2}{*}{Put Object Cabinet}
& $\pi_0$      & 68\% & 68\% & \textbf{72\%} \\
& $\pi_0$-FAST & 29\% & \textbf{32\%} & 28\% \\
\midrule
\multirow{2}{*}{Handover Block}
& $\pi_0$      & 44\% & 62\% & \textbf{63\%} \\
& $\pi_0$-FAST & 15\% & 16\% & \textbf{20\%} \\
\midrule
Avg. & Overall & 45.0\% & \textbf{51.5\%} & 51.0\% \\
\bottomrule
\end{tabular}%
}
\end{table}

\subsection{Ablation on Corrective Recovery Strategies}

We further compare CARE with a rollback-based recovery strategy on simulation
tasks. Rollback uses the same 3D monitor to detect deviations, restores execution
to the final state of the previous atomic stage, and then resumes nominal
execution without issuing an explicit corrective instruction for the current
failure. All compared variants use the standard simulation data budget: 150
nominal expert demonstrations and 50 additional trajectories per error type.
CARE uses corrective trajectories, while the baseline and rollback variants use
matched nominal atomic-stage trajectories.

As shown in Table~\ref{tab:rollback_vs_care}, rollback provides only limited
improvement. Returning to an earlier state can reduce accumulated error but does
not teach the policy how to act in the current failure state. CARE instead
combines explicit corrective supervision with atomic corrective execution and
achieves substantially larger gains.

\begin{table}[t]
\centering
\caption{Comparison of rollback-based recovery and CARE on three simulation tasks
using the $\pi_0$ backbone.}
\label{tab:rollback_vs_care}
\footnotesize
\setlength{\tabcolsep}{6pt}
\renewcommand{\arraystretch}{1.1}
\resizebox{\columnwidth}{!}{%
\begin{tabular}{lccc}
\toprule
Task & $\pi_0$ & $\pi_0$ (Rollback) & $\pi_0$ (CARE) \\
\midrule
Lift Pot          & 57\% & 67\% & \textbf{91\%} \\
Place Dual Shoes  & 14\% & 18\% & \textbf{40\%} \\
Stack Cubes Three & 48\% & 52\% & \textbf{61\%} \\
\midrule
Average            & 39.7\% & 45.7\% & \textbf{64.0\%} \\
\bottomrule
\end{tabular}%
}
\end{table}

\subsection{Accuracy of the 3D Monitor in Instruction Judgement}

We additionally evaluate the 3D monitor on triggered observations from a broader
set of simulation and real-world tasks. A monitoring decision is counted as
incorrect for either a false positive (a correct execution is judged to require
correction) or a false negative (a required correction is missed). 
The three simulation and two real-world tasks that overlap
with the main-paper monitoring ablation use exactly the same
reported accuracies.The additional tasks extend the analysis to object
diversity and real-world dual-arm manipulation.

\begin{table}[t]
\centering
\caption{Accuracy of the skill-level 3D monitor on simulation
and real-world tasks. Results for LP, PDS, HB, PTT, and STB
are identical to those in the main-paper monitoring ablation.}
\label{tab:3d_monitor_acc}
\footnotesize
\setlength{\tabcolsep}{7pt}
\renewcommand{\arraystretch}{1.05}
\resizebox{\columnwidth}{!}{%
\begin{tabular}{lc|lc}
\toprule
Task & Acc. & Task & Acc. \\
\midrule
Lift Pot             & 98\% & Place Bread Skillet & 92\% \\
Place Dual Shoes     & 95\% & Handover Block      & 96\% \\
Dump Bin Bigbin      & 94\% & Put Object Cabinet  & 96\% \\
Stamp Seal           & 96\% & Stack Two Cubes     & 90\% \\
Pass Two Toys        & 86\% & Stack Two Bowls     & 94\% \\
Place Two Cubes      & 88\% &                     &      \\
\midrule
\multicolumn{3}{l}{Average over all listed tasks} & 93.2\% \\
\bottomrule
\end{tabular}%
}
\end{table}

The monitor remains accurate despite substantial object variation. In
\textit{Put Object Cabinet}, which contains many object categories, the same
skill-level predicate templates achieve 96\% monitoring accuracy without a
task-specific learned classifier.

\section{Training and Evaluation Protocols}
\label{sec:training}

\subsection{Data budgets and VLA training parameters}

For standard simulation tasks in RoboTwin 2.0~\cite{chen2025robotwin} and
RoboFactory~\cite{qin2025robofactory}, each method uses 150 nominal expert
demonstrations and 50 additional trajectories per error type. CARE uses
experience-guided corrective trajectories for the additional data, whereas the
matched baseline uses nominal atomic-stage trajectories of the same scale.
The experience-guided distributions are estimated from 100 preliminary rollouts.

For the two VLA backbones used by CARE, $\pi_0$ and $\pi_0$-FAST, both the vanilla
and CARE variants are fine-tuned with the same optimization settings: 40{,}000
gradient steps, batch size 32, and action horizon 50 on two NVIDIA A800 SXM4
80GB GPUs. The backbone-specific learning-rate settings are given in
Tables~\ref{tab:train_config_pi0fast} and~\ref{tab:train_config_pi0}. ACT,
DP, DP3, and RDT-1B are trained using their model-specific benchmark recipes;
their configurations are listed for reproducibility in
Tables~\ref{tab:train_config_dp}--\ref{tab:train_config_rdt}.

For FSR-Bench, the data and optimization protocol is intentionally unified across
backbones: each recovery type uses 50 uniformly sampled corrective trajectories,
and all variants are trained for 20{,}000 steps with batch size 32 and action
horizon 50. CARE and its corresponding vanilla backbone use identical
hyperparameters so that the comparison isolates atomic corrective execution.

\begin{table}[t]
\centering
\caption{Unified FSR-Bench training protocol used by RDT-1B, $\pi_0$-FAST,
and $\pi_0$ variants. Backbone-specific optimizer settings follow the
corresponding model implementation.}
\label{tab:train_config_fsr}
\footnotesize
\setlength{\tabcolsep}{5pt}
\renewcommand{\arraystretch}{1.08}
\begin{tabular}{ll}
\toprule
Hyperparameter & Value \\
\midrule
Corrective trajectories / recovery type & 50 (uniformly sampled) \\
Training steps & 20{,}000 \\
Batch size & 32 \\
Action horizon & 50 \\
Baseline/CARE optimization & Identical within each backbone \\
\bottomrule
\end{tabular}
\end{table}

\subsection{Real-world setup and evaluation}

The real-world platform is a dual-arm LeRobot SO-101 system with 12 DoF in total
(6 per arm), one global RGB camera, and one wrist RGB camera per arm; all cameras
operate at 30\,Hz. For each real-world task, we collect 50 nominal demonstrations
and 20 additional trajectories per error type. CARE uses corrective
demonstrations for the additional data, while the matched baseline follows the
same data-budget protocol. We use 100 preliminary executions per task to estimate
the empirical failure distribution and evaluate standard task success over 100
trials per task.

The VLA policy is served from an A800 server during robot execution. SAM
3~\cite{carion2025sam} and DA3~\cite{lin2025depth} run locally on an RTX 4090 to
provide masks and depth for geometric monitoring. Separating policy inference
from perception keeps the event-triggered monitor responsive during deployment.

\subsection{Evaluation coverage and action vocabulary}

We evaluate seven hard bimanual tasks from RoboTwin 2.0, three multi-arm
collaboration tasks from RoboFactory, five FSR-Bench recovery tasks, and four
real-world dual-arm tasks. RoboTwin 2.0 uses the \texttt{demo\_randomized}
setting for both training and evaluation. FSR-Bench uses 100 trials per task--regime pair, while the standard
real-world tasks use 100 evaluation trials per task.

Across tasks, the VLM planner uses a shared atomic-action vocabulary, with
object-specific arguments inserted into the language instruction. Table
\ref{tab:action_vocab} lists representative action types and instruction forms.

\begin{table}[t]
\centering
\small
\caption{Unified action vocabulary after merging object-specific variations.}
\label{tab:action_vocab}
\setlength{\tabcolsep}{4pt}
\renewcommand{\arraystretch}{1.1}
\begin{tabular}{p{0.26\columnwidth} p{0.66\columnwidth}}
\toprule
Action Type & Example Instruction \\
\midrule
Wait & Wait for the other robots. \\
Grasp & Grasp the camera / bread / drawer bar / object. \\
Lift & Lift the object / camera / meat. \\
Place & Place the object / on the target / into the cabinet. \\
Re-grasp & Re-grasp the object. \\
Re-place & Re-place the object. \\
Adjust & Adjust the grasping / placement / press position. \\
Handover & Handover the object. \\
Press & Press the camera shutter button. \\
Move & Move the pot to the middle. \\
Pull & Pull the drawer. \\
Dump & Dump the dustbin. \\
Rotate & Rotate it back to the upright orientation. \\
Remove obstacle & Remove the obstacle. \\
Step back and wait & Step back and wait for the other robots. / Open the gripper and retreat. \\
\bottomrule
\end{tabular}
\end{table}

\begin{table}[t]
\centering
\caption{Training configuration of $\pi_0$-FAST in RoboTwin and RoboFactory.}
\label{tab:train_config_pi0fast}
\footnotesize
\setlength{\tabcolsep}{4pt}
\renewcommand{\arraystretch}{1.08}
\resizebox{\columnwidth}{!}{%
\begin{tabular}{ll}
\toprule
Hyperparameter & Value \\
\midrule
Model & $\pi_0$-FAST \\
Initialization & Base checkpoint initialization \\
Action horizon & 50 \\
Maximum token length & 280 \\
Batch size & 32 \\
Training steps & 40{,}000 \\
Peak learning rate & $5\times10^{-5}$ \\
Warmup steps & 200 \\
Decay steps & 1400 \\
Hardware & 2 $\times$ NVIDIA A800 GPUs \\
\bottomrule
\end{tabular}%
}
\end{table}

\begin{table}[t]
\centering
\caption{Training configuration of $\pi_0$ in RoboTwin and RoboFactory.}
\label{tab:train_config_pi0}
\footnotesize
\setlength{\tabcolsep}{4pt}
\renewcommand{\arraystretch}{1.08}
\resizebox{\columnwidth}{!}{%
\begin{tabular}{ll}
\toprule
Hyperparameter & Value \\
\midrule
Model & $\pi_0$ \\
Initialization & Base checkpoint initialization \\
Action horizon & 50 \\
Maximum token length & 180 \\
Batch size & 32 \\
Training steps & 40{,}000 \\
Peak learning rate & $3\times10^{-5}$ \\
Warmup steps & 300 \\
Decay steps & 10{,}000 \\
Final learning rate & $1\times10^{-6}$ \\
Hardware & 2 $\times$ NVIDIA A800 GPUs \\
\bottomrule
\end{tabular}%
}
\end{table}

\begin{table}[t]
\centering
\caption{Training configuration of DP in RoboTwin and RoboFactory.}
\label{tab:train_config_dp}
\footnotesize
\setlength{\tabcolsep}{4pt}
\renewcommand{\arraystretch}{1.08}
\resizebox{\columnwidth}{!}{%
\begin{tabular}{ll}
\toprule
Hyperparameter & Value \\
\midrule
Model & DP \\
Policy architecture & Diffusion U-Net image policy \\
Observation encoder & Multi-image ResNet-18 encoder \\
Horizon & 8 \\
Observation steps & 3 \\
Action steps & 6 \\
Diffusion steps (train / infer) & 100 / 100 \\
Batch size & 512 \\
Optimizer & AdamW \\
Learning rate & $1\times10^{-4}$ \\
Weight decay & $1\times10^{-6}$ \\
LR scheduler & Cosine \\
Warmup steps & 500 \\
Training epochs & 600 \\
Hardware & 1 $\times$ GPU (cuda:0) \\
\bottomrule
\end{tabular}%
}
\end{table}

\begin{table}[t]
\centering
\caption{Training configuration of DP3 in RoboTwin.}
\label{tab:train_config_dp3}
\footnotesize
\setlength{\tabcolsep}{4pt}
\renewcommand{\arraystretch}{1.08}
\resizebox{\columnwidth}{!}{%
\begin{tabular}{ll}
\toprule
Hyperparameter & Value \\
\midrule
Model & DP3 \\
Observation encoder & PointNet encoder \\
Horizon & 8 \\
Observation steps & 3 \\
Action steps & 6 \\
Crop shape & $80 \times 80$ \\
Diffusion steps (train / infer) & 100 / 10 \\
Noise scheduler & DDIM \\
Batch size & 3096 \\
Optimizer & AdamW \\
Learning rate & $1\times10^{-4}$ \\
Weight decay & $1\times10^{-6}$ \\
LR scheduler & Cosine \\
Warmup steps & 500 \\
Training epochs & 3000 \\
Hardware & 1 $\times$ GPU (cuda:0) \\
\bottomrule
\end{tabular}%
}
\end{table}

\begin{table}[t]
\centering
\caption{Training configuration of ACT in RoboTwin.}
\label{tab:train_config_act}
\footnotesize
\setlength{\tabcolsep}{4pt}
\renewcommand{\arraystretch}{1.08}
\resizebox{\columnwidth}{!}{%
\begin{tabular}{ll}
\toprule
Hyperparameter & Value \\
\midrule
Model & ACT \\
KL weight & 10 \\
Chunk size & 50 \\
Hidden dimension & 512 \\
Feedforward dimension & 3200 \\
Batch size & 64 \\
Learning rate & $1\times10^{-5}$ \\
Training epochs & 6000 \\
\bottomrule
\end{tabular}%
}
\end{table}

\begin{table}[t]
\centering
\caption{Training configuration of RDT-1B in RoboTwin.}
\label{tab:train_config_rdt}
\footnotesize
\setlength{\tabcolsep}{4pt}
\renewcommand{\arraystretch}{1.08}
\resizebox{\columnwidth}{!}{%
\begin{tabular}{ll}
\toprule
Hyperparameter & Value \\
\midrule
Model & RDT-1B \\
Train batch size & 32 \\
Sample batch size & 64 \\
Max training steps & 20000 \\
Learning rate & $1\times10^{-4}$ \\
Dataloader workers & 8 \\
State noise SNR & 40 \\
Gradient accumulation steps & 1 \\
Hardware & 1 $\times$ GPU (cuda:0) \\
\bottomrule
\end{tabular}%
}
\end{table}

\section{Visualization}
\label{sec:visualization}

\begin{figure*}[t]
    \centering
    \includegraphics[width=\textwidth]{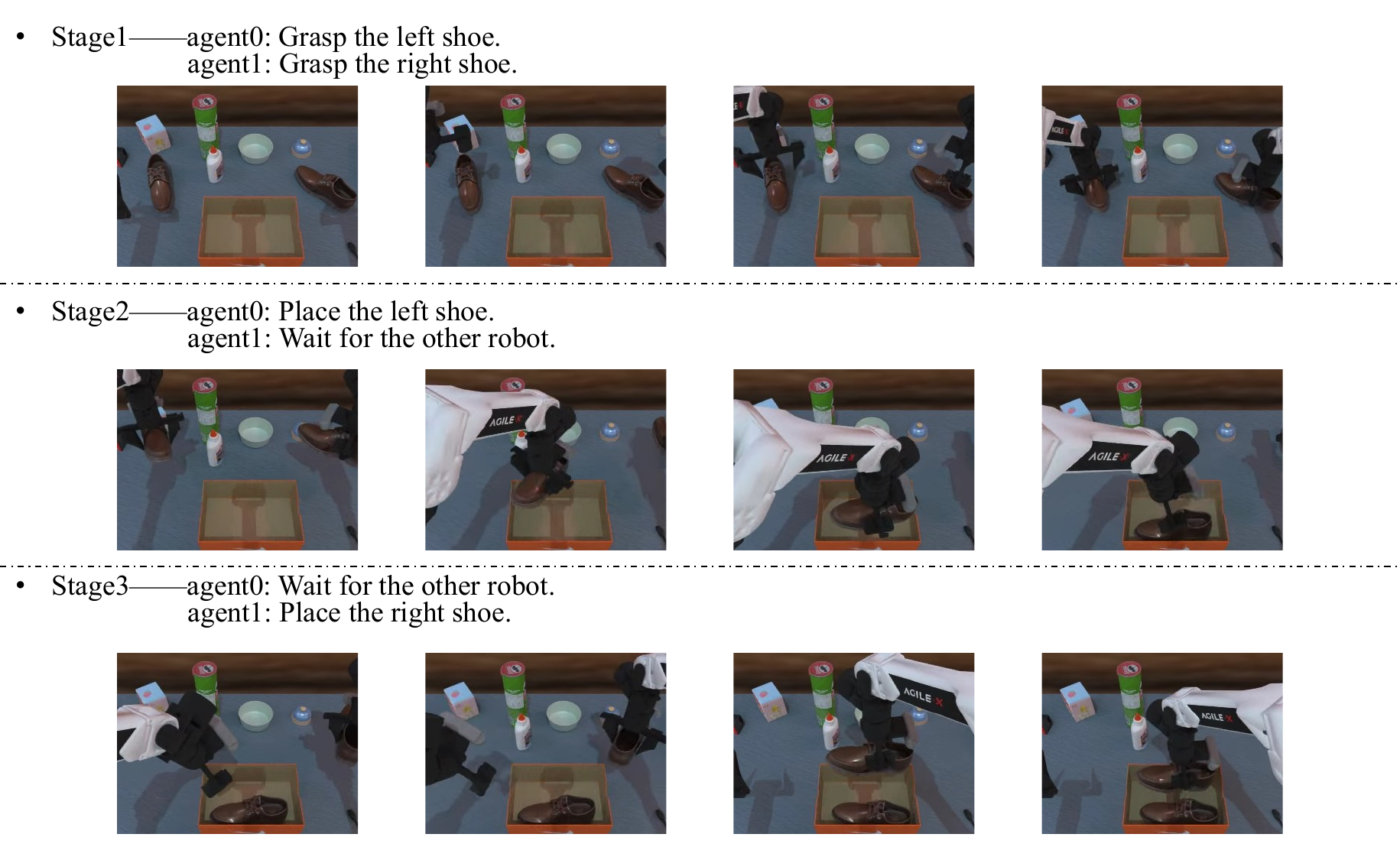}
    \caption{Visualization of the nominal execution for \textit{Place Dual Shoes} (Part 1/4). We show the full process of the three nominal atomic actions in the task.}
    \label{fig:nominal_shoes}
\end{figure*}

\begin{figure*}[t]
    \centering
    \includegraphics[width=\textwidth]{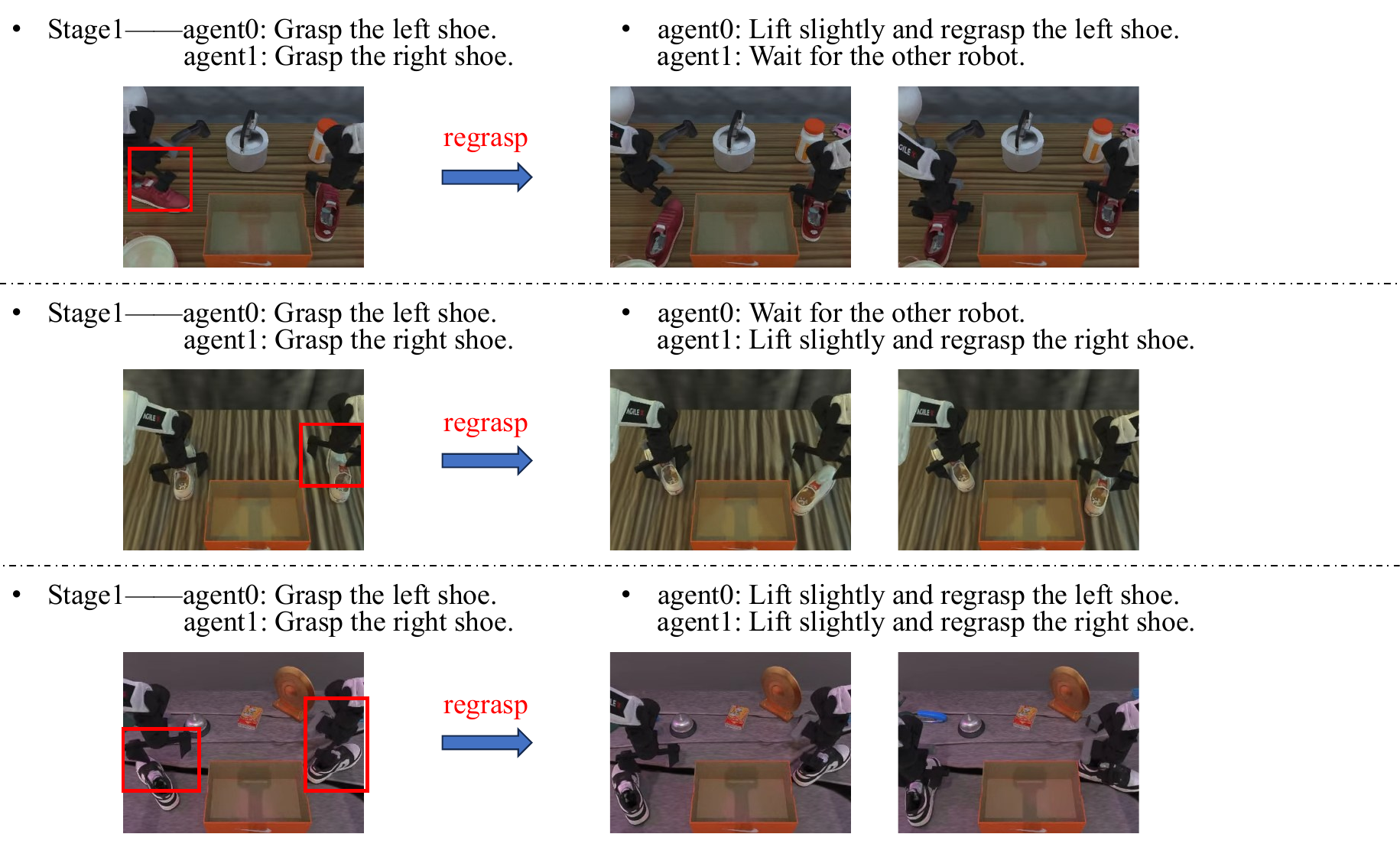}
    \caption{Visualization of the corrective execution for \textit{Place Dual Shoes} at Stage 1 (Part 2/4). The figure shows the correction process for the first stage of the task.}
    \label{fig:shoes_stage1}
\end{figure*}

We visualize the nominal execution and corrective execution cases for
\textit{Place Dual Shoes}. The nominal trajectory is shown in
Fig.~\ref{fig:nominal_shoes}, while the corrective processes for Stages 1--3 are
shown in Figs.~\ref{fig:shoes_stage1}--\ref{fig:shoes_stage3}. Additional
qualitative results for the remaining tasks are provided in the supplementary
demo video.

\begin{figure}[t]
    \centering
    \includegraphics[width=\columnwidth]{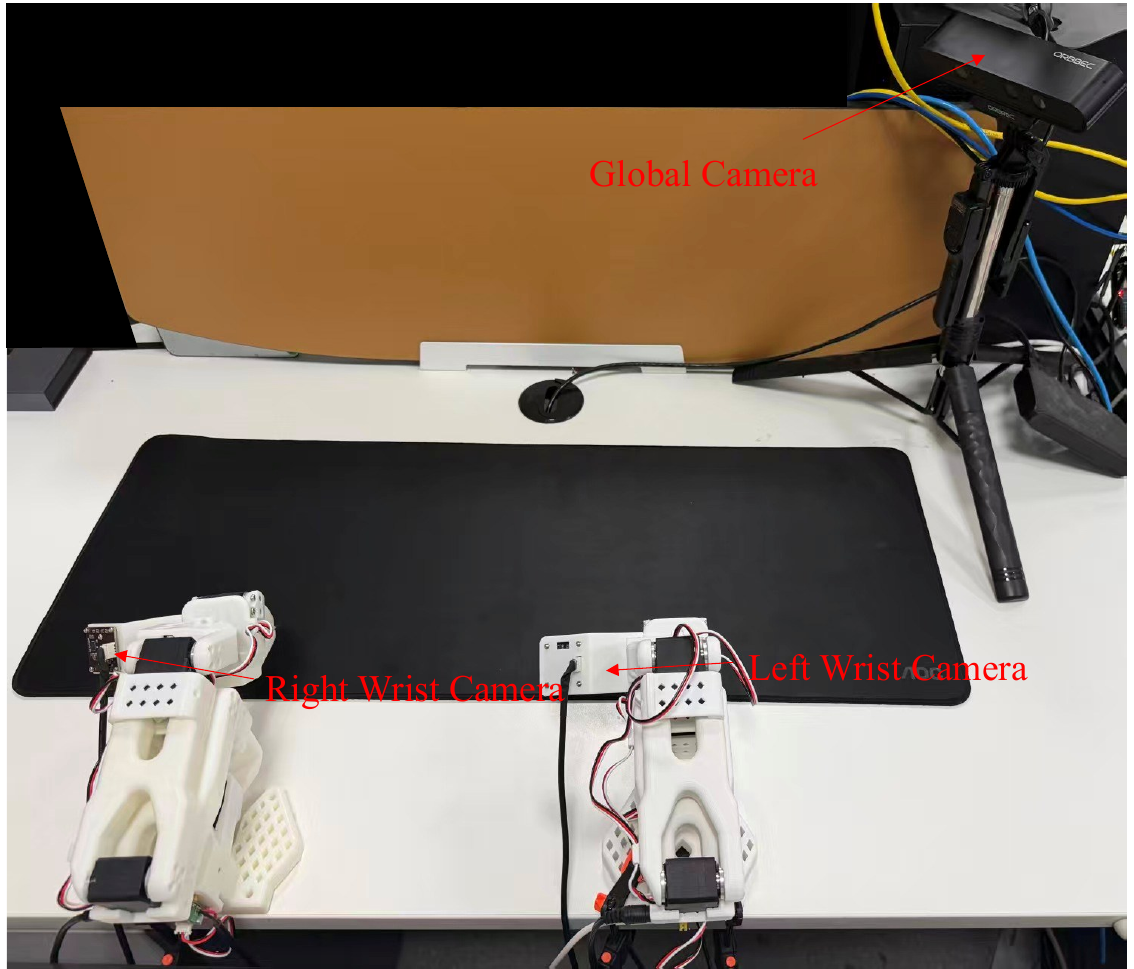}
    \caption{Real-world setup based on the dual-arm LeRobot SO-101 platform.}
    \label{fig:so101_setup}
\end{figure}

\begin{figure*}[t]
    \centering
    \includegraphics[width=\textwidth]{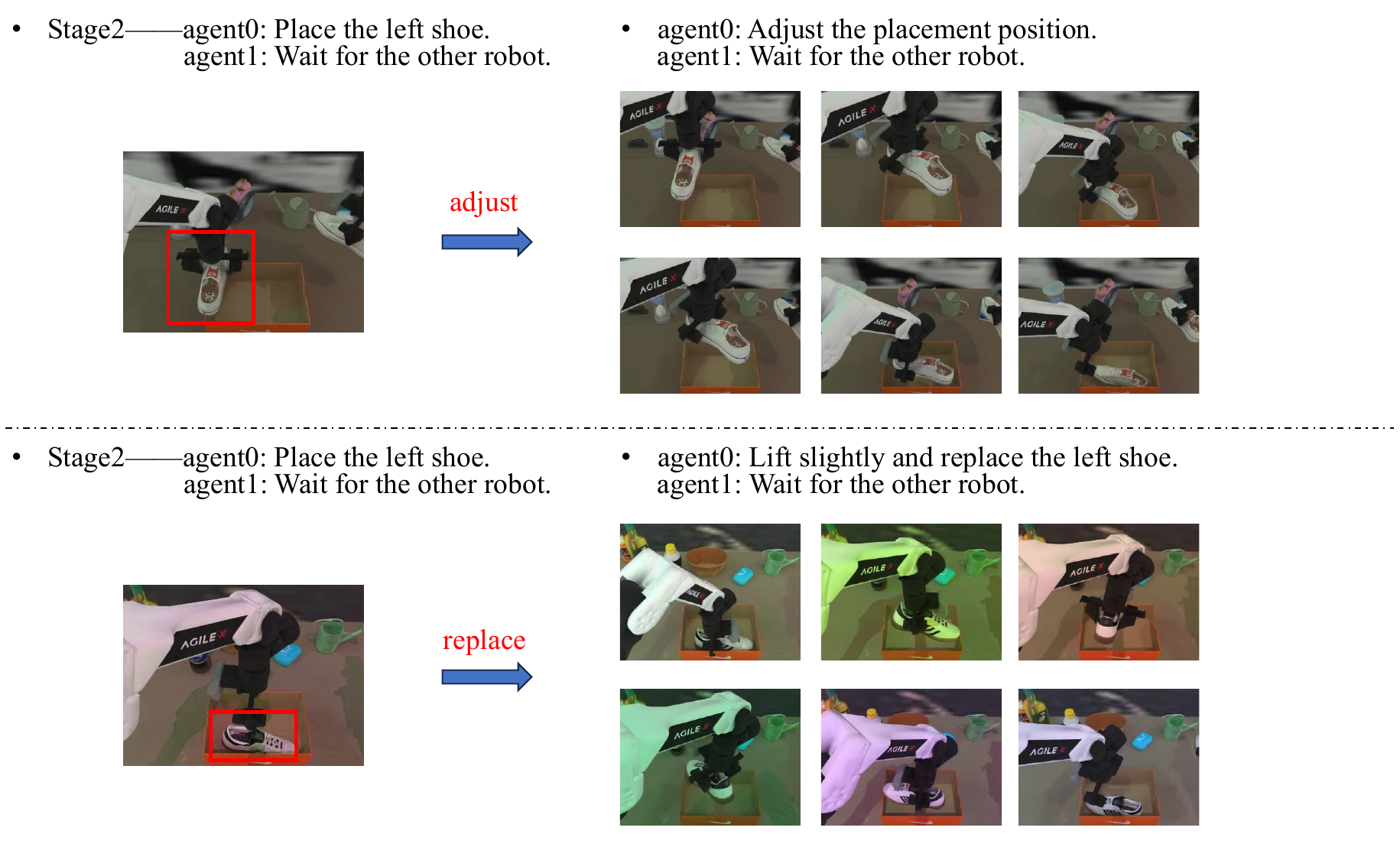}
    \caption{Visualization of the corrective execution for \textit{Place Dual Shoes} at Stage 2 (Part 3/4). We show the correction process corresponding to the second atomic stage of the task.}
    \label{fig:shoes_stage2}
\end{figure*}

\begin{figure*}[t]
    \centering
    \includegraphics[width=\textwidth]{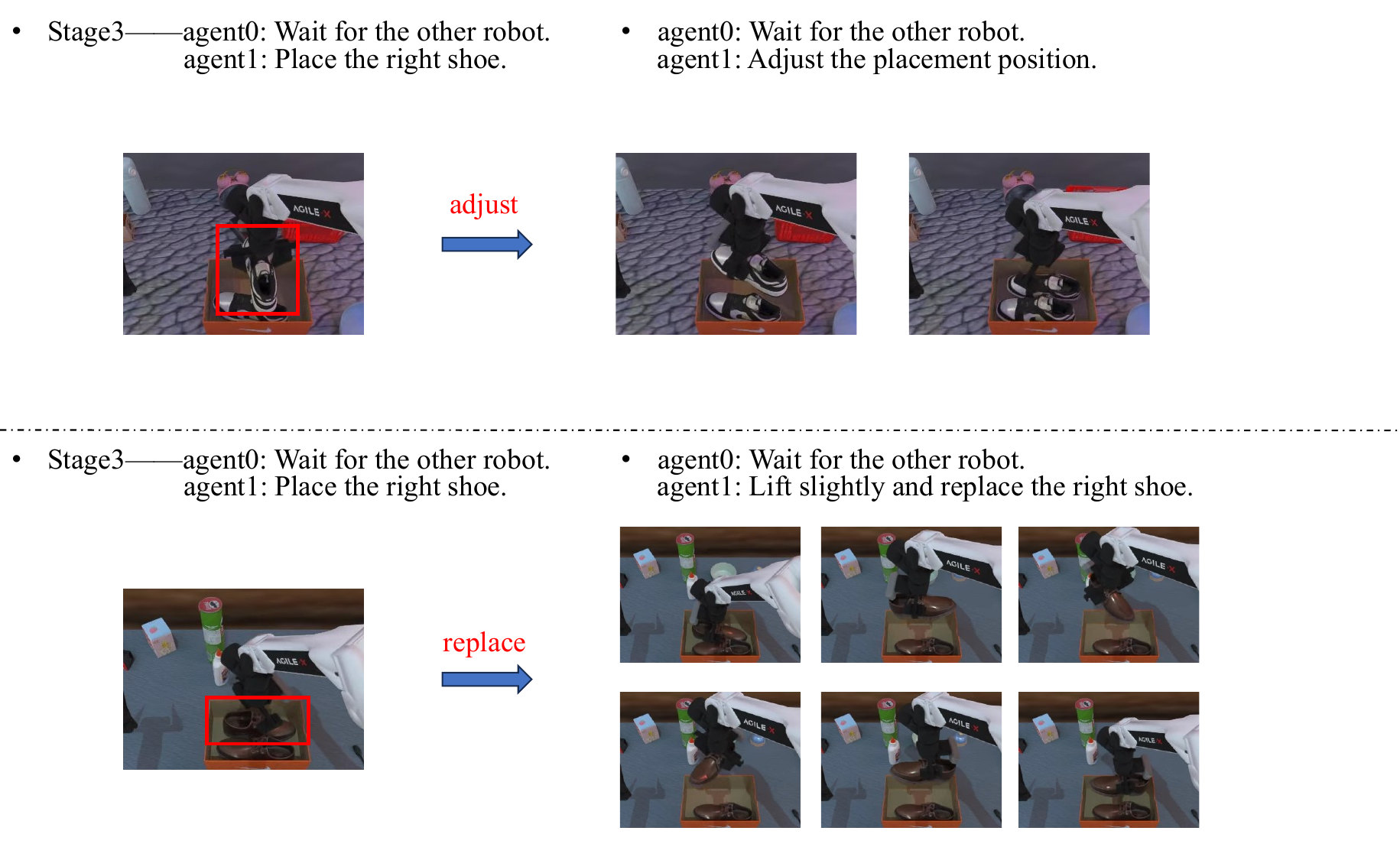}
    \caption{Visualization of the corrective execution for \textit{Place Dual Shoes} at Stage 3 (Part 4/4). We show the correction process corresponding to the third atomic stage of the task.}
    \label{fig:shoes_stage3}
\end{figure*}

\clearpage
\clearpage

\clearpage
\end{document}